%% file: main.tex
\documentclass{article}
\usepackage{amssymb}

\usepackage{times} %
\usepackage{graphicx}
\usepackage{amsmath} 
\usepackage{bbm}
\usepackage{makecell}
\usepackage{multirow}
\usepackage{booktabs}
\usepackage{wrapfig}
\usepackage{here}
\usepackage{enumitem}

\usepackage[T1]{fontenc}
\usepackage{xcolor}
\usepackage[table]{xcolor}
\usepackage{pifont}%

\usepackage{caption}
\usepackage{bm}
\usepackage{float}
\usepackage{makecell}
\usepackage{xspace}
\usepackage{calc}
\newcommand{\isaccepted}{1}
\usepackage[final]{corl_2026} %

\definecolor{orange}{HTML}{ED7046}
\definecolor{sky}{HTML}{087190}
\definecolor{light_mintgreen}{HTML}{E6EFE7}
\definecolor{mintgreen}{HTML}{BBDAC2}
\definecolor{citeblue}{HTML}{00075F}

\newcommand{\proposed}{\textcolor{black}{Flow as Flow}\xspace}
\newcommand{\flow}{\textcolor{black}{robot flow}\xspace}
\newcommand{\flows}{\textcolor{black}{robot flows}\xspace}

\newcommand{\firstmodule}{\textcolor{black}{the Flow Generation module}\xspace}
\newcommand{\secondmodule}{\textcolor{black}{the Action Generation module}\xspace}

\newcommand{\figref}{Fig.~\ref}
\newcommand{\tabref}{Table~\ref}
\newcommand{\ceqref}{Eq.~\eqref}
\newcommand{\eg}{e.g.,\xspace}

\title{Flow as Flow: Modeling Robot Velocity Fields \\ 
as Probability Velocity Fields \\for Flow-Based Object Manipulation}

\AtBeginDocument{
  \hypersetup{
    pdftitle={Flow as Flow: Modeling Robot Velocity Fields as Probability Velocity Fields for Flow-Based Object Manipulation},
    pdfauthor={Koki Seno, Daichi Yashima, Yusuke Takagi, Kento Tokura, Komei Sugiura}
  }
}

\author{
    \textbf{Koki~Seno} \, 
  \textbf{Daichi~Yashima} \, 
  \textbf{Yusuke~Takagi} \\ \,
  \textbf{Kento~Tokura} \, \textbf{Komei~Sugiura} \\
  \\[-0.5em]
  Keio University, Japan \\
  \\[-0.5em]
 \url{https://snkk62.github.io/flow-as-flow-page/}
}

\begin{document}
\maketitle

\begin{abstract}
Cross-embodiment data have become central to training robotic foundation models.
To leverage such heterogeneous data, we focus on flow-based object manipulation, where \textit{robot flows} (robot velocity fields) serve as embodiment-agnostic motion representations.
Previous studies do not formulate robot flows as dense velocity fields, but as displacements of sparse keypoints, 
even though dense velocity fields better match the continuous-time nature of motions.
To address this, we propose \proposed, a framework that models \flows as probability flows based on a flow matching formulation.
By naturally modeling such velocity fields within this formulation, our method achieves efficient and high-quality robot flow generation.
Across standard benchmarks, our method outperforms representative baseline methods on standard metrics, while achieving approximately 24$\times$ faster generation than standard flow matching.
Furthermore, through real-world experiments evaluating 9 methods with 260 trials per method across 13 manipulation tasks, we show that our method achieves a higher average success rate than the baseline methods.

\end{abstract}

\begin{quote}
\end{quote}

\setlength{\abovedisplayskip}{8pt} 
\setlength{\belowdisplayskip}{8pt}

\input{tab/1-10-eyecatch}

\input{section1}
\input{section2}

\input{section4-main}

\input{section6-main}

\input{section7}

\section*{Acknowledgments}
This work was partially supported by JST Moonshot and JSPS Fellows Grant Number JP26KJ1969.

\bibliography{reference}  %

\clearpage
\newcommand{\maincite}[1]{[\textcolor{citeblue}{#1}]}
{ \huge \textbf{Appendix}}
\setcounter{section}{0}   %
\renewcommand{\thesection}{\Alph{section}}  %

\input{supplymentaly}

\clearpage
\makeatletter
\let\oldthebibliography\thebibliography
\let\endoldthebibliography\endthebibliography

\renewenvironment{thebibliography}[1]{%
  \oldthebibliography{#1}%
  \setcounter{NAT@ctr}{67}%
}{%
  \endoldthebibliography
}
\makeatother

\end{document}

%% file: tab/1-10-eyecatch.tex
\begin{figure}[h]
    \centering
    \includegraphics[width=1.0\linewidth]{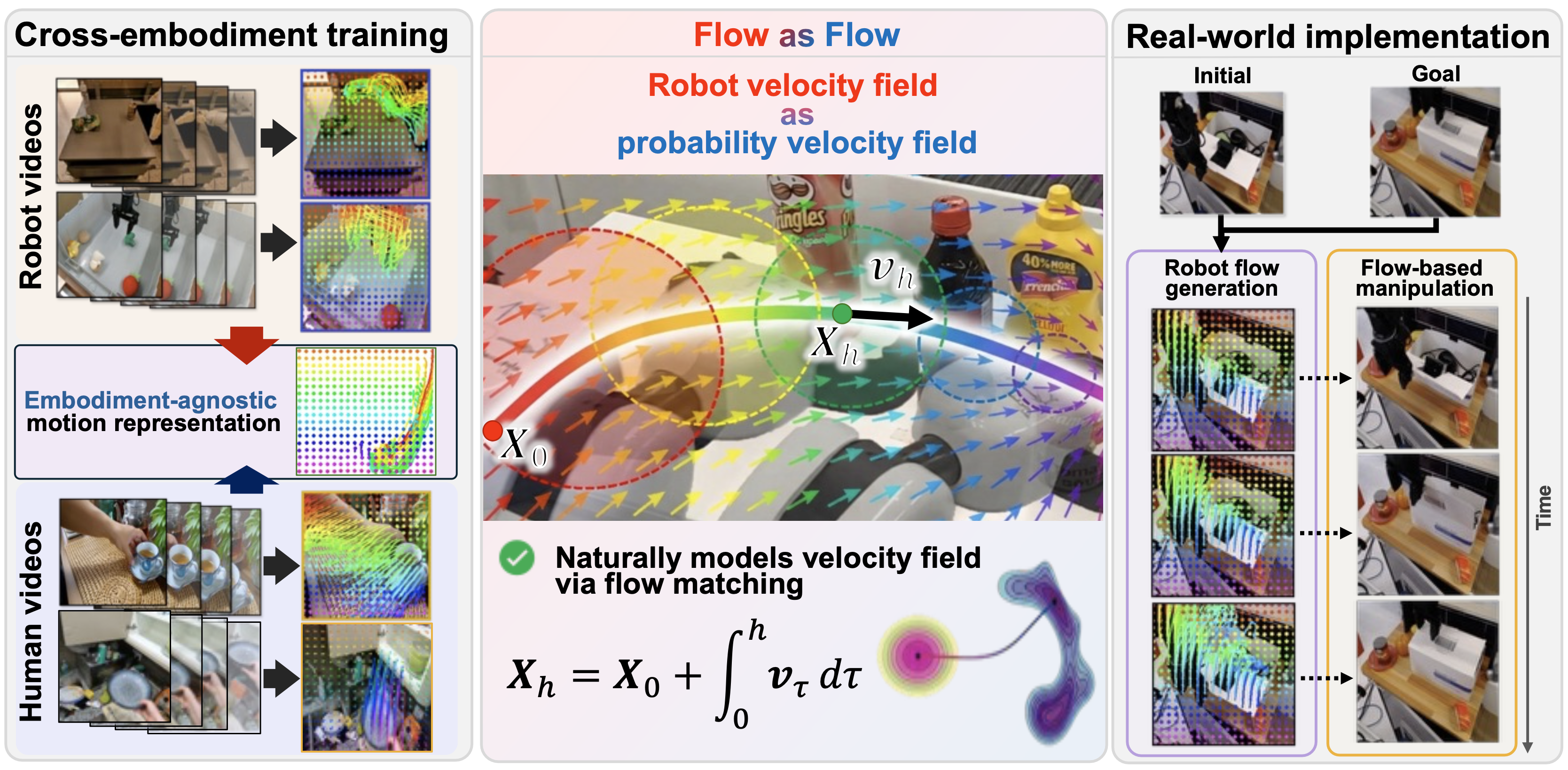}
    \caption{
    \small
    \textbf{Overview of our framework.} We use cross-embodiment data for training, which include human demonstrations as an additional embodiment. Our framework, \proposed{}, models \textit{robot flows} (robot velocity fields) as \textit{probability flows} using flow matching. At test time, it generates a \flow conditioned on an initial image and a goal image.
    The robot then executes object manipulation based on the generated flow.
    }
    \label{fig:eye-catch}
\end{figure}

%% file: section1.tex
\vspace{-1mm}
\section{Introduction} \label{intro}
\vspace{-2mm}

Labor shortages pose a significant challenge and increase demand for robotic systems that support human workers and improve operational efficiency in a range of environments, including factories, warehouses, and shopping malls.
Accordingly, robotic foundation models have become an active area of research across diverse tasks and environments~\cite{pi0-black25rss, pi05-black25corl, gr00t-bjorck25, giga-brain-ye25}.
A key factor behind this progress is large-scale pretraining, which makes training data central to model capabilities.
Thus, a growing body of studies has demonstrated the benefits of increasing the scale of training data~\cite{pi0-black25rss, oxe-neill24icra, scaling-law-il-lin25iclr, vlaser-yang26iclr}.
However, a bottleneck remains in scaling data collection, which is caused by labor-intensive teleoperation using real robots, as existing methods typically require substantial teleoperation for each embodiment.

To tackle this bottleneck, 
we focus on flow-based object manipulation, where \textit{\flows} (robot velocity fields) serve as motion-centric representations for action execution.
These representations can be readily extracted from visual observations collected across diverse embodiments, including both robots and humans, without requiring additional processing to reconcile embodiment-specific discrepancies.
Therefore, the flow-based approach enables scalable policy learning from low-cost and widely available video data, which substantially reduces the dependence on robot-specific data.

Some recent efforts have explored flow-based object manipulation~\cite{track2act-bharadhwaj24eccv, im2flow2act-xu24corl, atm-wen23rss}; however, they do not formulate robot flows as dense velocity fields, but as displacements of sparse keypoints, even though such velocity fields better match the continuous-time nature of motions.
These approaches can be seen as predicting velocity fields under the strong assumption that the resulting keypoint displacements are uniquely determined. However, this assumption is rarely satisfied in practice.
Consequently, the generated \flows remain only rough approximations of their underlying velocity fields.
Therefore, these works do not account for this approximation, which can become problematic in real-world robot implementation.

To address this limitation, we propose \proposed, a framework that models robot velocity fields as probability velocity fields; it achieves efficient and high-quality \flow generation.
It is derived from the perspective that robot motions are inherently physical processes governed by temporal evolution and therefore can be formulated as velocity fields.
Thus, our method leverages a flow matching framework, which models probability velocity fields, for modeling robot motions in a physically consistent manner.
The key difference between our method and existing methods is that \proposed models robot flows as continuous-time probability velocity fields using flow matching.

\vspace{-1mm}

\figref{fig:eye-catch} shows an overview of our framework.
For training, we leverage videos from cross-embodiment datasets covering multiple robot embodiments and humans.
At deployment, the model predicts a \flow that represents a task-relevant motion (\eg closing the lid of a cardboard box) conditioned on an initial image and a goal image.
Subsequently, the robot executes object manipulation conditioned on the generated flow to achieve the object poses specified by the goal image.

\vspace{-1mm}
The core contributions of this study are as follows:
\vspace{-2mm}
\begin{itemize}[leftmargin=2.8em, labelsep=0.6em]
    \setlength{\parskip}{0.3mm} %
    \setlength{\itemsep}{0.3mm} %
    \item We propose \proposed, a framework that models image-space robot velocity fields as continuous-time probability velocity fields, which achieves the efficient and high-quality generation of \flows{}.
    \item In the flow generation task, our method outperforms baseline methods on Fractal~\cite{rt1-brohan23rss}, Bridge V2~\cite{bridgev2-walke23corl}, DROID~\cite{droid-khazatsky24rss}
     and Fanuc Manipulation~\cite{fanuc-zhu23} datasets that include zero-shot settings, while achieving approximately 24$\times$ faster generation than standard flow matching.
    \item Through real-world experiments, we show that our method achieves a higher average success rate than representative baseline methods, which highlights the effectiveness of our framework and the utility of \flows for conditioning manipulation policies.
\end{itemize}

%% file: section2.tex
\vspace{-1mm}
\section{Related Work}
\vspace{-1mm}

\textbf{Flow Matching.}
Diffusion models and flow matching
have demonstrated strong capabilities across a wide range of tasks, including image generation (\eg{}~\cite{sd-rombach22cvpr, sd3-esser24icml, dit-peebles23iccv}) and object segmentation (\eg{}~\cite{ov-seg-diffusion-li23-iccv, seg-diffusion-iioka23iros}).
In robot learning, recent work has increasingly leveraged these expressive frameworks to generate actions~\cite{pi0-black25rss, flowpolicy-zhang-aaai25, reinflow-zhang25neurips, dp-chi23rss}.
Despite their strong generation capability, a key limitation is that these frameworks typically require many sampling steps at inference time, which leads to a bottleneck in generation speed.
To mitigate this issue, autoregressive diffusion and flow matching models generate an underlying probabilistic path in an autoregressive manner, which enables fast generation with only a small number of sampling steps while preserving the expressive formulation of diffusion models and flow matching~\cite{ar-diffusion-sun25cvpr, flowar-ren25icml, hiflow-yashima26}.
Along similar lines, 
Streaming Flow Policy is most closely related to our work.
It incorporates negative feedback typical in control theory, which is proportional to the trajectory error relative to a reference trajectory, and improves robustness under out-of-distribution conditions~\cite{stabilization-block23neurips, sfp-jiang25corl}.

\vspace{-1mm}
\textbf{Flow-Based Object Manipulation.}
Flow-based manipulation models commonly use \flows as motion representations that bridge vision and action, and they achieve strong results 
under limited real robot data settings
~\cite{atm-wen23rss, generalflow-yuan25corl, vidbot-chen25cvpr, flowretrieval-lin25corl, skil-wang25rss}.
A large body of work focuses on 2D \flows, which provide a simple and effective interface for learning robot motions from videos~\cite{robotap-vecerik24icra, track2act-bharadhwaj24eccv, flip-gao25iclr, im2flow2act-xu24corl, hinflow-zheng26iclr}.
Some methods use hand- or object-centric flows, whereas others use flow-conditioned world models~\cite{im2flow2act-xu24corl, flip-gao25iclr, atm-wen23rss, vidbot-chen25cvpr}.
Other approaches also extend flows to 3D or 6-DoF trajectories using depth estimation~\cite{vidbot-chen25cvpr, flowbot3d-eisner22rss, egoscaler-yoshida25cvpr, dev-vla-from-ego-yoshida26icra}.
Unlike these methods that predict keypoint coordinates, \proposed predicts robot velocity fields directly via flow matching, which naturally models velocity fields.

\vspace{-1mm}
\textbf{Learning from Cross-Embodiment Data.}
To address the scarcity of real-world robot data, learning from large-scale human videos has been actively explored~\cite{univla-bu25rss, gr00t-bjorck25, track2act-bharadhwaj24eccv, im2flow2act-xu24corl, egoscaler-yoshida25cvpr}.
Prior work explored various approaches, including pretraining visual features~\cite{vip-ma23iclr, r3m-nair23corl}, learning reward functions~\cite{in-the-wild-reward-chen21rss, xirl-zakka22corl}, hand pose tracking~\cite{videodex-shaw23corl}, affordance extraction~\cite{affordance-from-human-video-bahl23cvpr, human-hands-as-probes-goyal22cvpr, interaction-hotspots-liu22cvpr}, and domain translation~\cite{context-translation-liu18icra, mirage-chen20rss, rl-with-videos-schmeckpeper21corl}.
Among these, a growing line of work emphasizes scaling data by leveraging large-scale human video datasets~\cite{ego4d-grauman22cvpr, epickitchens-damen18eccv, sthsthv2-goyal17iccv, lapa-ye25iclr}.
A central challenge in leveraging web-scale videos, particularly human videos, is the lack of action labels that are typically required for policy learning~\cite{lapa-ye25iclr, univla-bu25rss, universal-policies-du23neurips}. 
To bridge this gap, many methods extract motion-centric representations from videos to construct pseudo action labels~\cite{gr00t-bjorck25, lapa-ye25iclr, univla-bu25rss, generalflow-yuan25corl, atm-wen23rss}.
These representations enable learning motion priors from web-scale videos for training manipulation policies with limited real robot demonstrations.

%% file: section4-main.tex
\vspace{-3mm}
\section{Approach} \label{method}
\vspace{-2mm}
\subsection{Preliminary: Flow Matching}
\vspace{-2mm}
Flow matching~\cite{flowmatching-lipman23iclr} is a generative modeling framework that learns a time-dependent probability velocity field that transports samples from a simple prior distribution $p_{\mathrm{prior}}$ (\eg a Gaussian distribution) to a target distribution $p_{\mathrm{data}}$~\cite{flowmatching-lipman23iclr}.
In this framework, the generation process starts from a sample $\bm{x}_0 \sim p_{\mathrm{prior}}$ and the intermediate state $\bm{x}_t$ evolves over time $t \in [0,1]$ governed by the ordinary differential equation (ODE):
{
\setlength{\abovedisplayskip}{6pt}
\setlength{\belowdisplayskip}{6pt}
\begin{align}
\frac{d \bm{x}_t}{dt} = \mathbf{v}(\bm{x}_t,t),
\end{align}
}where $\mathbf{v}$ denotes the probability velocity field.  
Although directly estimating the target distribution is natural, it is typically intractable; this formulation instead models the probability velocity field in a generation space, enabling training via simple regression on the velocity field.
The generated sample at $t=1$ is given by
{
\setlength{\abovedisplayskip}{6pt}
\setlength{\belowdisplayskip}{5pt}
\begin{align}
\bm{x}_1 = \bm{x}_0 + \int_0^1 \bm{\mathbf{v}}(\bm{x}_\tau,\tau)\,d\tau.
\end{align}
}During training, a velocity model $\mathbf{v}_\psi$ parametrized by $\psi$ is trained via regression on $\mathbf{v}(\bm{x}_t,t)$ using the following objective:
{
\setlength{\abovedisplayskip}{5pt}
\setlength{\belowdisplayskip}{5pt}
\begin{align}
\mathcal{L}_\text{FM}=
\mathbb{E}_{\bm{x}_t,t}
\big[\|\mathbf{v}_\psi(\bm{x}_t,t)-\mathbf{v}(\bm{x}_t,t)\|^2\big].
\end{align}
}

\input{tab/4-4-models}
\input{section3}
\vspace{-5mm}
\subsection{\proposed}
\vspace{-2mm}
We propose \proposed{}, a framework for object manipulation via \flow generation that fundamentally extends flow-based policies~\cite{track2act-bharadhwaj24eccv, flip-gao25iclr, im2flow2act-xu24corl}.
This framework is grounded in the view that robot motions are inherently time-evolving physical processes and can be naturally formulated as velocity fields.
Thus, \proposed formulates \textit{robot flows} as \textit{probability flows} in generation space.
This formulation leverages an expressive flow matching framework to represent robot motions as continuous-time velocity fields.

The novelty of our framework is that it models image-space robot velocity fields as probability velocity fields, achieving efficient and high-quality generation of \flows.
\proposed can be incorporated into flow-based object manipulation frameworks by modifying only the training objective, without changing the model architecture or introducing additional modules.
Indeed, our framework can be directly integrated into Track2Act~\cite{track2act-bharadhwaj24eccv}, FLIP~\cite{flip-gao25iclr}, or Im2Flow2Act~\cite{im2flow2act-xu24corl} without any structural changes (see~\tabref{tab:flow_quantitative}).

We initialize $N$ (\eg $10 \times 10$) points uniformly on $\mathcal{I}$. We obtain their future positions by integrating the velocity fields predicted by a flow generation model $\bm{v}_\theta$.
Let $\bm{\Xi}_h \in \mathbb{R}^{N\times2}$ denote the ground-truth point coordinates at timestep $h$ by arranging the coordinates of the $N$ points into an $N\times2$ matrix.
The corresponding velocities are denoted by $\dot{\bm{\Xi}}_h \in \mathbb{R}^{N\times2}$.
We construct target velocity fields in a typical stabilizing feedback form with proportional feedback on a tracking error:
{
\setlength{\abovedisplayskip}{8pt}
\setlength{\belowdisplayskip}{8pt}
\begin{align}
\bm{v}(\bm{\Xi}_h, \bm{X}, h) = \dot{\bm{\Xi}}_h - k(\bm{X} - \bm{\Xi}_h),
\label{eq:velocity}
\end{align}
}where $\bm{X} \sim \mathcal{N}\!\left(\bm{\Xi}_h,\ \sigma_0^2 e^{-2kh}\bm{I}\right)$ with a small constant $\sigma_0 > 0$. 
The second term 
is a stabilization term with the coefficient $k$, which enhances robustness to out-of-distribution samples~\cite{sfp-jiang25corl, stabilization-block23neurips}.

We train $\bm{v}_\theta$ using the conditional flow matching (CFM) loss~\cite{flowmatching-lipman23iclr} defined as follows:
{
\setlength{\abovedisplayskip}{4pt}
\setlength{\belowdisplayskip}{4pt}
\begin{align}
\mathcal{L}_\text{CFM} = \mathbb{E}_{\bm{\Xi}_h, h, \bm{X}} \left[ \left\lVert  \bm{v}_{\theta}\left(\bm{X}, h, \mathcal{I}, \mathcal{G}, \bm{\Xi}_{0:h-1} \right) - \bm{v}\left(\bm{\Xi}_h, \bm{X}, h \right)\right\rVert^2 \right],
\end{align}
}where $\bm{\Xi}_{0:h-1} \triangleq [\bm{\Xi}_0, \dots, \bm{\Xi}_{h-1}]$.
The coordinates at $h$, denoted by $\bm{X}_h$, are obtained 
as follows:
{
\setlength{\abovedisplayskip}{4pt}
\setlength{\belowdisplayskip}{4pt}
\begin{align} \label{eq:generation}
\bm{X}_h = \bm{X}_0 + \int_0^h \bm{v}_\theta \left( \bm{X}_\tau, \tau , \mathcal{I}, \mathcal{G}, \mathcal{X}_{< \tau} \right) d\tau,
\end{align}
}where $\mathcal{X}_{<\tau}$ denotes the history of the point coordinates of previous timesteps.
As shown in ~\ceqref{eq:generation}, the model predicts the velocity field at each timestep conditioned on past coordinates, which enables fast generation in a manner inspired by autoregressive diffusion and flow matching methods (e.g., ~\cite{flowar-ren25icml}).
Finally, we obtain $\mathcal{X}_{0:H-1}$ by concatenating $\bm{X}_h$ from $h=0$ to $h=H-1$, where $H$ denotes the sequence length of the generated flow and $\mathcal{X}_{0:H-1} \triangleq [\mathbf{X}_0, \dots, \mathbf{X}_{H-1}]$.

\vspace{-1mm}
\subsection{Model Architecture
\label{sec:model_architecture}}
\vspace{-1mm}
\figref{fig:model} shows the architecture of our method.
It consists of two modules: \firstmodule and \secondmodule.
We leverage Flow as Flow in \firstmodule{}.

\noindent \textbf{The Flow Generation module.} \proposed does not depend on a particular model architecture for $\bm{v}_\theta$; we implement $\bm{v}_\theta$ based on the Diffusion Transformer (DiT)~\cite{dit-peebles23iccv}, following previous work~\cite{track2act-bharadhwaj24eccv}.
In our method, each DiT block is modulated via adaLN-Zero~\cite{dit-peebles23iccv} using a shared conditioning vector constructed from $\mathcal{I}$, $\mathcal{G}$, and $h$.
Specifically, $\mathcal{I}$ and $\mathcal{G}$ are embedded using ResNet-18~\cite{resnet-he16cvpr}, whereas $h$ is embedded using sinusoidal positional encoding.
The conditioning vector is obtained by summing these embeddings.
At each step $h$, the input sequence $\mathcal{X}_{0:h} \in \mathbb{R}^{(h+1) \times N \times 2}$ is constructed in an autoregressive manner by concatenating the past point coordinates with the current point coordinates.
The coordinates of the $N$ points in each step of the sequence are flattened into vectors in $\mathbb{R}^{2N}$ and treated as tokens. Then $\mathcal{X}_{0:h}$ is processed by DiT, and the output token at the latest step represents the predicted velocity.

\noindent \textbf{The Action Generation module.}
For object manipulation, \secondmodule~predicts the end-effector pose conditioned on $\mathcal{X}_{0:H-1}$ generated by \firstmodule.
At each timestep $t$, \secondmodule outputs $\bm{a}_t$ given $\mathcal{O}_t$, $\mathcal{X}_{0:H-1}$, and the robot state $\bm{s}_t$.
We implement the module based on Diffusion Policy (DP)~\cite{dp-chi23rss}. Specifically, we adopt a DiT-based architecture~\cite{dit-peebles23iccv} rather than the U-Net architecture used in the original paper~\cite{dp-chi23rss} because previous work demonstrated that DiT-based DP variants outperform the original U-Net-based DP~\cite{dit-block-policy-dasari25icra, mdt-reuss24rss}.
Similar to \firstmodule, $\mathcal{O}_t$ is encoded using ResNet-18, and conditioning signals are injected via adaLN-Zero.

Following previous work~\cite{flowpolicy-zhang-aaai25}, we adopt flow matching, which provides a strong alternative to diffusion for action generation.
We train the velocity model $\bm{u}_\phi$ using the CFM loss:
{
\setlength{\abovedisplayskip}{4pt}
\setlength{\belowdisplayskip}{4pt}
\begin{align}
\mathcal{L}_\text{act} = \mathbb{E}_{t,r,\bm{\varepsilon},\bm{a}_t}
\left[ \left\|  \bm{u}_\phi\!\left(\bm{a}_t^{r}, r, \mathcal{O}_t, \mathcal{X}_{0:H-1}, \bm{s}_t \right) - (\bm{a}_t - \bm{\varepsilon}) \right\|^2 \right],
\end{align}
}where $r \sim \mathcal{U}(0, 1)$, $\bm{\varepsilon} \sim \mathcal{N}(\mathbf{0}, \mathbf{I})$, and $\bm{a}_t^r$ denotes the noisy action at generation timestep $r$. 
Here, $\bm{a}_t^{r}$ is obtained by interpolating $\bm{a}_t$ with $\bm{\varepsilon}$ along the standard optimal transport path~\cite{flowmatching-lipman23iclr}.
For inference, $\bm{a}_t$ is generated by integrating the ODE defined by $\bm{u}_\phi$ from $r=0$ to $r=1$.

%% file: tab/4-4-models.tex
\begin{figure*}[t]
    \centering
    \includegraphics[width=1.0\linewidth]{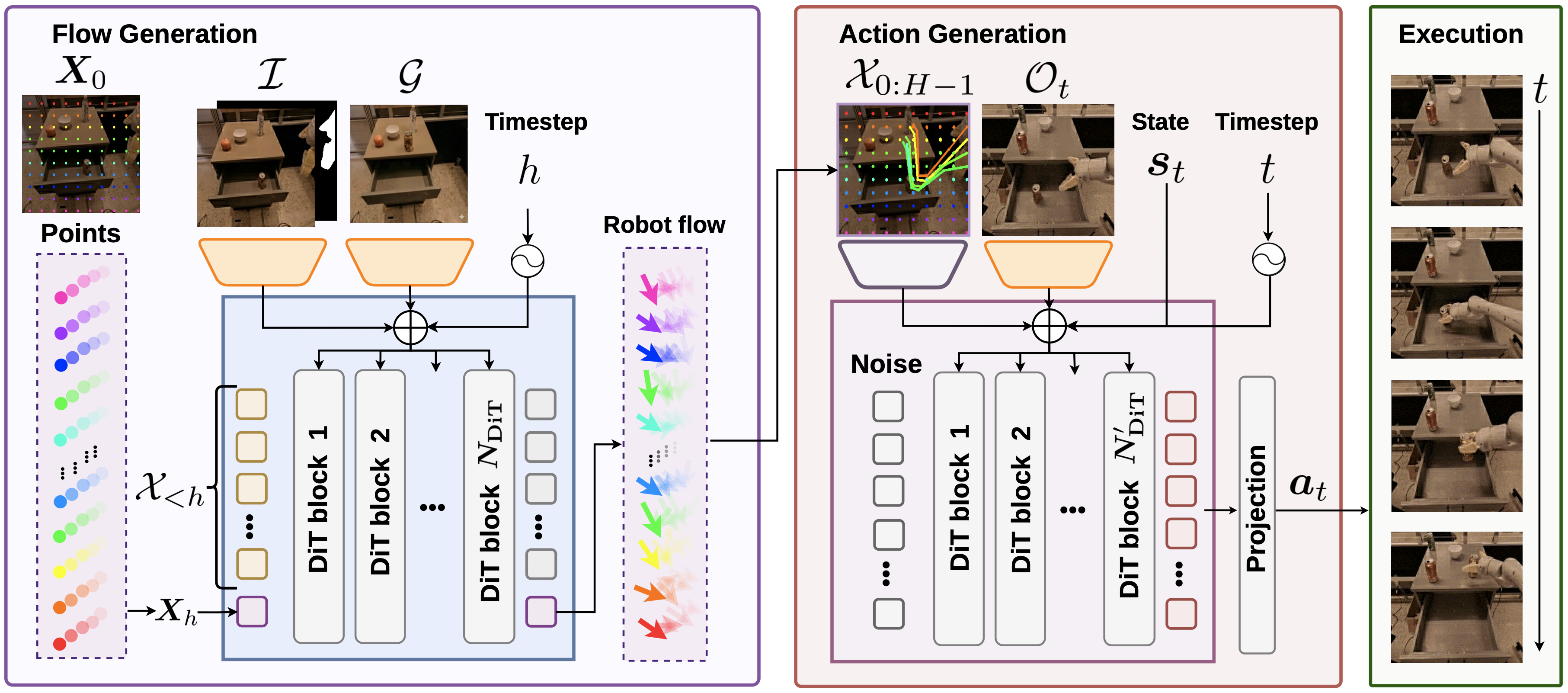}
    \vspace{-5mm}
    \caption{\textbf{Architecture of our method.} The Flow Generation module generates $\mathcal{X}_{0:H-1}$, and the Action Generation module predicts $\bm{a}_t$ based on it.
$\bm{X}_h$ and $\mathcal{X}_{<h}$ denote the point coordinates at timestep $h$ and their history, respectively. $N_{\mathrm{DiT}}$ and $N^\prime_{\mathrm{DiT}}$ are the numbers of the Diffusion Transformer blocks in \firstmodule and \secondmodule{}, respectively.}
    \vspace{-6mm}
    \label{fig:model}
\end{figure*}

%% file: section3.tex
\vspace{-1mm}
\subsection{Problem Statement}
\vspace{-1mm}
In this study, we focus on flow-based object manipulation conditioned on a goal image. 
Goal-conditioned object manipulation tasks are practical in settings where the desired final state of the environment can be specified explicitly, such as a room tidying task (\eg{}~\cite{tidybot-wu23auto-robo}).
In this scenario, an image of an organized environment can be captured in advance and used as the goal image.
Based on the image, the robot manipulates objects to achieve the specified goal state.

\vspace{-1mm}

The task is decomposed into two subtasks: \flow generation and object manipulation.
In the \flow generation subtask, given an initial image $\mathcal{I}$ and a goal image $\mathcal{G}$, the model is expected to generate a \flow{} aligned with the ground-truth end-effector motion.
Subsequently, in the object manipulation subtask, at each timestep $t$, given the current observation $\mathcal{O}_t$ and the generated flow, the model outputs the end-effector pose $\bm{a}_t$.
In this study, we focus on 2D end-effector flows under a fixed camera setup.
We assume that the end-effector is visible in both the initial and goal images.

%% file: section6-main.tex
\vspace{-3mm}
\section{
    Experiments
    \label{sec:experiments}
}
\vspace{-2mm}
 \subsection{
    Experimental Setup 
    \label{sec:experimental_settings}
}
\vspace{-1mm}
\input{section5}

 \textbf{Baselines.}
We used FLIP~\cite{flip-gao25iclr}, Im2Flow2Act~\cite{im2flow2act-xu24corl}, and GigaWorld-0-Video~\cite{giga-world-0-ye25} as language-conditioned baseline methods, and Track2Act~\cite{track2act-bharadhwaj24eccv} as a goal-conditioned baseline method.
We included GigaWorld-0-Video, a world model specialized for robotic environments, to assess the applicability of a world model to the flow generation task.
For GigaWorld-0-Video, a video was generated from $\mathcal{I}$ and a language input, and the \flow was subsequently obtained using CoTracker3~\cite{cotracker3-karaev25iccv}.
Implementation details are provided in Appendix~\ref{app:implementation}.

\textbf{Evaluation Metrics.} 
We used Average Displacement Error (ADE) and Final Displacement Error (FDE) as standard evaluation metrics~\cite{ egoscaler-yoshida25cvpr, flip-gao25iclr}, together with Less Than Delta Ratio (LTDR)~\cite{track2act-bharadhwaj24eccv, flip-gao25iclr, cotracker-karaev24eccv, tapvid-doersch22neurips}.
ADE served as the primary metric.
For LTDR, at each step $h$ and threshold $m$, $\delta^h_m$ denotes the fraction of points for which the Euclidean distance between $\bm{X}_h$ and $\bm{\Xi}_h$ does not exceed $m$ pixels.
The final score was computed by averaging $\delta^h_m$ over thresholds and the flow horizon as:
{
\setlength{\abovedisplayskip}{4pt}
\setlength{\belowdisplayskip}{4pt}
\begin{align}
\text{LTDR} = \frac{1}{MH} \sum_{m=1}^{M} \sum_{h=0}^{H-1} \delta_m^h .
\end{align}
}
Further details of the evaluation metrics are provided in Appendix~\ref{app:evaluation}.

\vspace{-2mm}
\subsection{Quantitative Results
\label{sec:quantitative}
}
\tabref{tab:flow_quantitative} shows a quantitative comparison between our method and the baseline methods on the test sets of Fractal, Bridge V2, DROID-100 and Fanuc Manipulation.
The results on Fractal and Bridge V2 correspond to the in-domain settings, whereas the results on DROID-100 and Fanuc Manipulation correspond to the zero-shot settings.
The zero-shot settings involve datasets with substantially different robot embodiments and environments compared with those seen during training.
Zero-shot performance is essential for evaluating robustness to robot embodiments and environments that differ substantially from the training data.

\input{tab/6-1-quantitative}

\tabref{tab:flow_quantitative} shows that our method achieved the best ADE scores of $21.23$, $27.11$, $35.89$, and $22.46$ on Fractal, Bridge V2, DROID-100, and Fanuc Manipulation, respectively. 
Our method outperformed the best baseline scores of $37.14$, $47.29$, $40.73$ and $27.37$  by $15.91$, $20.18$, $4.84$, and $4.91$ points, respectively.
The performance differences between our method and the baseline methods were statistically significant ($p<0.01$).

To further examine the effectiveness of our framework in a language-conditioned setting, we incorporated Flow as Flow into FLIP and Im2Flow2Act.
In these settings, their flow generation modules were trained within our framework.
The results show that our framework improved the ADE scores of both methods across the benchmarks.
Notably, the ADE scores of Im2Flow2Act decreased from $37.14$, $51.48$, $44.14$, and $38.15$ to $33.21$, $42.96$, $38.87$, and $26.51$ on Fractal, Bridge V2, DROID-100, and Fanuc Manipulation, respectively.
These results suggest that our framework was effective across multiple model architectures.
Furthermore, our framework accelerated inference for each method.
In particular, compared with Track2Act, our method reduced the inference time from $1{,}430$ ms to $44$ ms, which resulted in approximately $33\times$ faster generation.

\vspace{-2mm}
\subsection{Qualitative Results}
\vspace{-1mm}
Fig.~\ref{fig:flow-qualitative} shows qualitative results of robot flow generation, comparing our method with a baseline method (Track2Act).
In the figure, Rows (i)--(iii) correspond to samples from Fractal, Bridge V2, and Fanuc Manipulation, respectively.
The two left columns represent $\mathcal{I}$ and $\mathcal{G}$.
The other columns show the robot flows of the ground truth, Track2Act, and our method, where the flows are overlaid on $\mathcal{I}$.
For our method, we also visualized predicted flows at intermediate timesteps.

Row (i) shows that the robot grasped the can inside the drawer and placed it onto the counter.
In this case, Track2Act generated the \flow toward an incorrect object, whereas our method generated the flow that targeted the correct object and moved toward the appropriate location.
Row (ii) shows that the robot moved the white object near the image center toward the bottom-right region.
The resulting Track2Act motion deviated from the task-required direction, although it targeted the correct object.
By contrast, our method successfully generated the flow grasping the target object and moving it in the correct direction.
Row (iii) 
corresponds to a zero-shot setting.
Track2Act failed to generate meaningful flows of the end-effectors, whereas our method generated flows that exhibited plausible end-effector motions.
These results indicate that our method generated appropriate \flows for end-effector motions under in-domain settings and maintained robust performance in zero-shot scenarios.

\input{tab/6-8-qualitative}

\vspace{-4mm}
\input{tab/6-10-ablation}

\vspace{-2mm}
\subsection{Ablation Study
\label{sec:ablation_study}
}
\vspace{-2mm}
\tabref{tab:flow-ablation} presents the results of an ablation study, where we evaluated the effectiveness of \proposed by replacing it with standard flow matching~\cite{flowmatching-lipman23iclr}. 
The results show that the replacement degraded the ADE scores by 7.99, 3.45, 4.72, and 6.16 points on Fractal, Bridge V2, DROID-100, and Fanuc Manipulation, respectively.
Moreover, inference became approximately 24$\times$ slower, requiring 1,062 ms per sample compared with 44 ms.
These results suggest the effectiveness of our framework in achieving both faster and higher-quality flow generation than standard flow matching.

\vspace{-2mm}
\section{Real-World Experiments}
\vspace{-2mm}
\input{section6-physical}

%% file: section5.tex
We leveraged diverse web videos from both human and robot manipulation domains to train \firstmodule.
Specifically, we used human video clips from Something-Something-v2~\cite{sthsthv2-goyal17iccv} and EPIC-KITCHENS~\cite{epickitchens-damen18eccv}.
We also used robot video clips from standard robot manipulation datasets Fractal~\cite{rt1-brohan23rss} and Bridge V2~\cite{bridgev2-walke23corl} for training.
For evaluation, we used Fractal and Bridge V2 as in-domain benchmarks, and further assessed zero-shot performance on DROID-100~\cite{droid-khazatsky24rss} and Fanuc Manipulation~\cite{fanuc-zhu23}, which are also standard benchmarks in robot learning~\cite{oxe-neill24icra, openvla-kim24corl, univla-bu25rss}.
DROID-100 is a subset of the DROID dataset. 
Because our flow-based formulation is embodiment-agnostic, we additionally incorporated large-scale human video datasets into our training data.
The dataset details are shown in Appendix~\ref{app:dataset}.
Following previous work~\cite{track2act-bharadhwaj24eccv}, we set $H = 8$.
We trained our method using two NVIDIA H200 SXM GPUs (VRAM 141 GB). We used a single H200 for inference.
The total training time was approximately 19 hours.

%% file: tab/6-1-quantitative.tex
\begin{table}[t]
  \centering
  \small
  \setlength{\tabcolsep}{4pt}
  \renewcommand{\arraystretch}{1.2}
  \caption{Quantitative comparison with the baseline methods, where `inf. speed' represents the inference speed. The best and second-best scores are shown in \textbf{bold} and \underline{underlined}, respectively.}
  \vspace{1mm}
 \resizebox{\linewidth}{!}{
  \setlength{\aboverulesep}{0pt}
  \setlength{\belowrulesep}{0pt}
  \begin{tabular}{
    l |
    c |
    c c c 
    c c c 
    c c c 
    c c c 
    c
  }
    \toprule
    \multirow{3}{*}{Method} & 
    \multirow{3}{*}{\makecell[c]{Flow-\\as Flow}} & %
    \multicolumn{6}{c}{In-domain} & \multicolumn{6}{c}{Zero-shot}  &
    \multirow{3}{*}{%
      \makecell[c]{Inf.\\speed $\downarrow$\\{[ms]}}
    } \\
    \cmidrule(lr){3-8}\cmidrule(lr){9-14} &
    &
    \multicolumn{3}{c}{Fractal~\cite{rt1-brohan23rss}} &
    \multicolumn{3}{c}{Bridge V2~\cite{bridgev2-walke23corl} } &
    \multicolumn{3}{c}{DROID-100~\cite{droid-khazatsky24rss}} &
    \multicolumn{3}{c}{Fanuc Manipulation~\cite{fanuc-zhu23}} & \\
    \cmidrule(lr){3-5}\cmidrule(lr){6-8}
    \cmidrule(lr){9-11}\cmidrule(lr){12-14} 
    & & {ADE $\downarrow$} & {FDE $\downarrow$} &{LTDR $\uparrow$[\%]} &
     {ADE $\downarrow$} & {FDE $\downarrow$} & {LTDR $\uparrow$[\%]} &
     {ADE $\downarrow$} & {FDE $\downarrow$} & {LTDR $\uparrow$[\%]} &
     {ADE $\downarrow$} & {FDE $\downarrow$} & {LTDR $\uparrow$[\%]} & {} \\ 
    \midrule

    \multicolumn{15}{l}{\textit{Language-conditioned}} \\
      
    \midrule
    
    FLIP~\cite{flip-gao25iclr}
      & 
      & 66.17 & 87.52 & 35.69 
      & 50.73 & 68.43 & 47.72
      & 43.10 & 49.10 & 54.87
      & 28.31 & 50.83 & \underline{72.17}
      & \underline{35} \\
      
    \rowcolor{light_mintgreen}FLIP
      & \checkmark
      & 38.77 & 57.41	& 58.11	
      & 48.34 & 66.31	& 49.26
      & \underline{38.54} & \underline{44.48}	& \underline{56.25}
      & 26.79 & 47.85	& 71.62
      & \textbf{17} \\
      
    Im2Flow2Act~\cite{im2flow2act-xu24corl}
      & 
      & 37.14 & 47.74 & 60.61
      & 51.48 & 70.93 & 47.97
      & 44.14 & 54.41 & 51.48
      & 38.15 & 64.18 & 59.25 
      & 5,580 \\
      
    \rowcolor{light_mintgreen}Im2Flow2Act
      & \checkmark
      & \underline{33.21} & \underline{46.83}	& \underline{64.25}
      & \underline{42.96} & \underline{60.93}	& \underline{54.00}
      & 38.87 & 45.25	& 56.07
      & \underline{26.51} & 48.48	& 71.75
      & 230 \\
      
    GigaWorld-0-Video~\cite{giga-world-0-ye25}
      & --
      & 74.00 & 95.23 & 32.46
      & 53.18 & 69.58 & 46.44
      & 42.75 & 47.96 & 53.91
      & 37.60 & 58.35 & 61.22 
      & 26,976 \\
      
    \midrule
      
    \multicolumn{15}{l}{\textit{Goal-conditioned}} \\
      
    \midrule
    
    Track2Act~\cite{track2act-bharadhwaj24eccv}
      & 
      & 64.32 & 86.62 & 42.00
      & 47.29 & 64.13 & 51.61
      & 40.73 & 47.43 & 54.29
      & 27.37 & \underline{47.17} & 70.99
      & 1,430 \\

    \rowcolor{mintgreen}\textbf{Ours}
      & \checkmark
      & \textbf{21.23} & \textbf{27.31} & \textbf{76.79}
      & \textbf{27.11} & \textbf{34.66} & \textbf{69.96}
      & \textbf{35.89} & \textbf{40.58} & \textbf{58.81}
      & \textbf{22.46} & \textbf{42.19} & \textbf{74.54} 
      & 44 \\

    \bottomrule
  \end{tabular}
  }
  \vspace{-5mm}
  \label{tab:flow_quantitative}
\end{table}

%% file: tab/6-8-qualitative.tex
\begin{figure*}[t]
    \centering
    \includegraphics[width=0.95\linewidth]{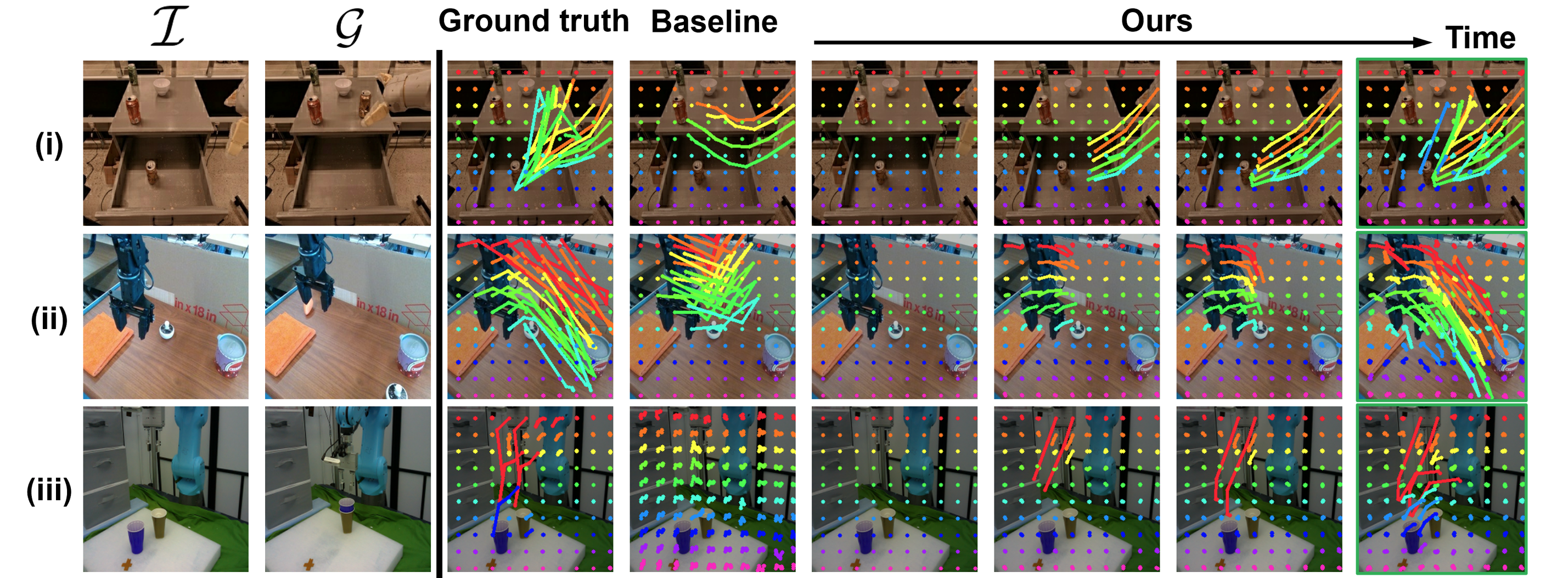}
    \vspace{-1mm}
    \caption{Qualitative comparison of robot flow generation between our method and a baseline method (Track2Act). Rows (i), (ii), and (iii)
    correspond to samples from Fractal, Bridge V2,
    and Fanuc Manipulation, respectively. The final prediction of our method is highlighted in green.
    }
    \vspace{-4mm}
    \label{fig:flow-qualitative}
\end{figure*}

%% file: tab/6-10-ablation.tex
\begin{table}[H]
  \centering
  \small
  \setlength{\tabcolsep}{4pt}
  \renewcommand{\arraystretch}{1.5}
  \caption{Ablation study on the framework for the Flow Generation module. Our framework is compared with standard flow matching~\cite{flowmatching-lipman23iclr}. `inf. speed' represents inference speed.}
  \vspace{1mm}
 \resizebox{\linewidth}{!}{
  \setlength{\aboverulesep}{0pt}
  \setlength{\belowrulesep}{0pt}
  \begin{tabular}{
    l |
    c c c
    c c c
    c c c
    c c c
    c
  }
    \toprule
    \multirow{2}{*}{Framework} & 
    \multicolumn{3}{c}{Fractal~\cite{rt1-brohan23rss}} &
    \multicolumn{3}{c}{Bridge V2~\cite{bridgev2-walke23corl} } &
    \multicolumn{3}{c}{DROID-100~\cite{droid-khazatsky24rss}} &
    \multicolumn{3}{c}{Fanuc Manipulation~\cite{fanuc-zhu23}} &
    \multirow[c]{2}{*}{%
      \makecell[c]{Inf. speed $\downarrow$\\{[ms]}}
    } \\
    \cmidrule(lr){2-4}\cmidrule(lr){5-7}
    \cmidrule(lr){8-10}\cmidrule(lr){11-13} 
    & {ADE $\downarrow$} & {FDE $\downarrow$} &{LTDR $\uparrow$[\%]} &
     {ADE $\downarrow$} & {FDE $\downarrow$} & {LTDR $\uparrow$[\%]} &
     {ADE $\downarrow$} & {FDE $\downarrow$} & {LTDR $\uparrow$[\%]} &
     {ADE $\downarrow$} & {FDE $\downarrow$} & {LTDR $\uparrow$[\%]} & {} \\ 
    \midrule

    Flow Matching
      & 29.22 & 38.30 & 70.12 
      & 30.56 & 38.76 & 67.32
      & 40.61 & 45.70 & 56.85
      & 28.62 & 51.94 & 71.34
      & 1,062 \\

    \textbf{Flow as Flow}
      & \textbf{21.23} & \textbf{27.31} & \textbf{76.79}
      & \textbf{27.11} & \textbf{34.66} & \textbf{69.96}
      & \textbf{35.89} & \textbf{40.58} & \textbf{58.81}
      & \textbf{22.46} & \textbf{42.19} & \textbf{74.54} 
      & \textbf{44} \\

    \bottomrule
  \end{tabular}
  }
  \vspace{-4mm}
  \label{tab:flow-ablation}
\end{table}

%% file: section6-physical.tex
\input{tab/6-physical-setup}

\subsection{Experimental Setup}
\vspace{-1mm}
We conducted real-world experiments to validate our method in downstream mobile manipulation tasks.
We evaluated the quality of robot flows in Sec.~\ref{sec:experiments}, whereas we discuss the downstream manipulation performance in this section.
We focused on mobile manipulation as a practical setting for robot use in workplace environments.
In the experiments, we used the Human Support Robot~\cite{hsr-yamamoto19robomech}, an 11-DoF mobile manipulator developed by Toyota Motor Corporation with cameras mounted on both the head and end-effector.
$\mathcal{O}_t$ was captured by the robot's head and wrist cameras, whereas $\mathcal{I}$ and $\mathcal{G}$ were captured by a fixed third-person webcam.
Before each trial, the objects were manually arranged in the desired final configuration to capture $\mathcal{G}$ and then rearranged into the initial configuration.
Fig.~\ref{fig:real-world-setup} shows representative examples of the mobile manipulation tasks performed in our experiments.
We evaluated our method on a diverse set of 13 mobile manipulation tasks.
For each task, we collected approximately 30 teleoperated demonstrations for training.
Detailed task descriptions are provided in Appendix~\ref{app:task}.

\textbf{Baselines}.
We used three language-conditioned and two goal-conditioned baselines.
FLIP~\cite{flip-gao25iclr}, Im2Flow2Act~\cite{im2flow2act-xu24corl}, Track2Act~\cite{track2act-bharadhwaj24eccv}, and our method are categorized as flow-based methods.
They first generated \flows and conditioned their manipulation policies on the generated flows.
To assess the benefit of this approach, we used DP-Lang and DP-Goal as additional baselines.
Neither was conditioned on robot flows, but on language instructions or goal images.
They were based on DP~\cite{dp-chi23rss}, a standard and competitive method that has demonstrated strong performance even with limited real-robot demonstrations~\cite{cdp-ma25corl}.
Consistent with the Action Generation module 
(Sec.~\ref{sec:model_architecture}),
we implemented them with DiT-based~\cite{dit-peebles23iccv} architectures.

\input{tab/6-physical}
\input{tab/6-physical-qualitative}

\vspace{-2mm}
\subsection{Quantitative Results}
\vspace{-2mm}
Table \ref{tab:real-world-results} shows the quantitative results of the real-world experiments.
For each method, we report per-task success rates and the average success rate across tasks.
Our method achieved an average success rate of 58\%, outperforming the strongest baseline method, Track2Act, by 10 points.
Similar to \tabref{tab:flow_quantitative}, the table shows that incorporating Flow as Flow into FLIP and Im2Flow2Act increased their average success rates from 28\% to 43\% and from 45\% to 55\%, respectively.
These results indicate that our framework was also effective in the downstream mobile manipulation tasks.

Additionally, the table shows that the flow-based methods generally achieved higher average success rates than both the DP variants.
These results highlight the advantage of \flows over language instructions and goal images for conditioning policies. A discussion is provided in Appendix~\ref{app:flow-effectiveness}.

\vspace{-2mm}
\subsection{Qualitative Results}
\vspace{-1mm}
\figref{fig:real-world-qualitative} shows qualitative results of our method in the real-world experiments.
\figref{fig:real-world-qualitative} (a) presents a \textbf{bussing table} case, where the robot grasped the apple on the table and moved it to the cardboard box.
Our method appropriately generated the robot flow that reflected both the target object and the destination, even though this cluttered configuration was not observed during training.
Conditioned on the flow, the policy successfully generated the motion.
\figref{fig:real-world-qualitative} (b) illustrates a \textbf{close laptop} case, involving the manipulation of the hinged screen.
Our method generated the flow appropriate for the laptop's opening angle, and the robot closed the laptop successfully.
Moreover,~\figref{fig:real-world-qualitative} (c) shows a \textbf{push chair} case in a human workspace, where the robot successfully pushed the chair under the table. 
This demonstrates that the generated robot flow captured motion associated with whole-body mobile-manipulator movement, not only end-effector motion.

%% file: tab/6-physical-setup.tex
\begin{figure*}[t]
    \centering
    \includegraphics[width=0.98\linewidth]{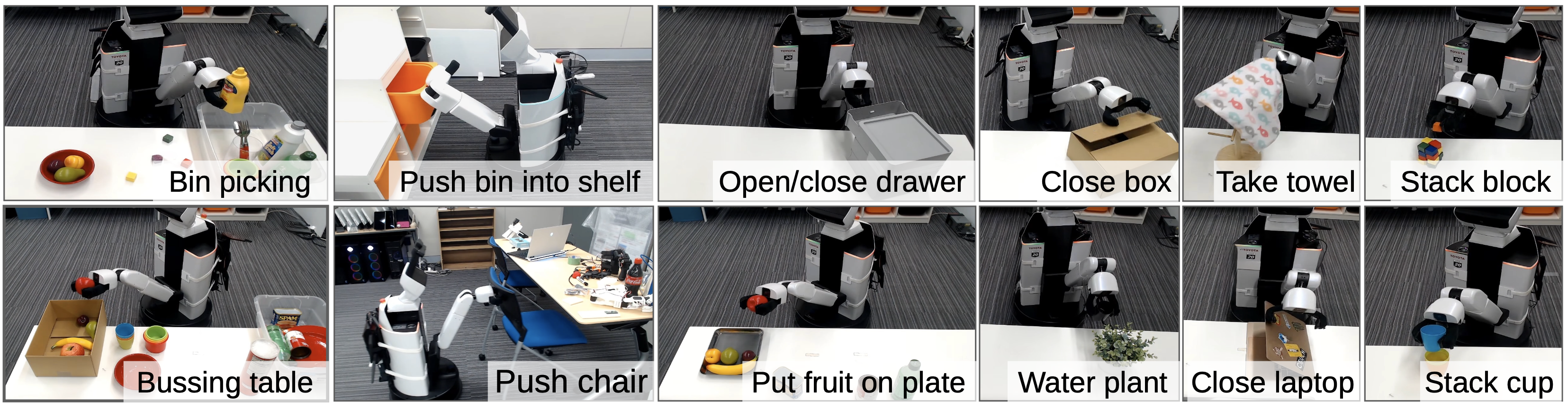}
    \vspace{-1mm}
    \caption{Representative examples from our real-world experiments. We conducted experiments on a broad set of 13 manipulation tasks. Each task involves mobile manipulation, which extends standard tabletop manipulation by requiring mobile base movement.}
    \vspace{-4.5mm}
    \label{fig:real-world-setup}
\end{figure*}

%% file: tab/6-physical.tex
\begin{table}[t]
  \centering
  \small
  \setlength{\tabcolsep}{4pt}
  \renewcommand{\arraystretch}{1.2}
  \caption{Quantitative results in real-world experiments. We conducted 260 trials for each method, with 20 trials for each task.
For the oracle method, the Action Generation module was conditioned on a robot flow extracted from the corresponding teleoperation videos using CoTracker3~\cite{cotracker3-karaev25iccv}. 
}
\vspace{1mm}

  \resizebox{\linewidth}{!}{
  \setlength{\aboverulesep}{0pt}
  \setlength{\belowrulesep}{0pt}
  \begin{tabular}{
    l |
    c
    c |
    c c c
    c c c
    c c c
    c c c
    c |
    c
  }
    \toprule
    \makecell[l]{\multicolumn{1}{c}{[\%]} \\ Method} 
    & \makecell[c]{Flow-\\based}
    & \makecell[c]{Flow-\\as Flow}
    & \makecell[c]{Push bin\\into shelf}
    & \makecell[c]{Push\\chair}
    & \makecell[c]{Close\\drawer}
    & \makecell[c]{Close\\box}
    & \makecell[c]{Take\\towel}
    & \makecell[c]{Bin\\picking}
    & \makecell[c]{Put fruit\\on plate}
    & \makecell[c]{Bussing\\table}
    & \makecell[c]{Close\\laptop}
    & \makecell[c]{Water\\plant}
    & \makecell[c]{Open\\drawer}
    & \makecell[c]{Stack\\cup}
    & \makecell[c]{Stack\\block}
    & Avg. \\
      
    \midrule
    \multicolumn{17}{l}{\textit{Language-conditioned}} \\
    \midrule

    DP-Lang
      & 
      & 
      & 75 & 75 & 65
      & \underline{65} & 35 & 35
      & 35 & 35 & 35
      & 5 & 20 & 5
      & 10
      & 38 \\

    FLIP~\cite{flip-gao25iclr}
      & \checkmark
      & 
      & 45 & 55 & 50
      & 60 & 50 & 25
      & 15 & 15 & 25
      & 5 & 10 & 10
      & 5
      & 28 \\

    \rowcolor{light_mintgreen}FLIP %
      & \checkmark
      & \checkmark
      & 55 & 55 & 70
      & \underline{65} & 55 & 45
      & \underline{60} & 45 & 40
      & \underline{25} & \underline{25} & 10
      & 10
      & 43 \\

    Im2Flow2Act~\cite{im2flow2act-xu24corl}
      & \checkmark
      &
      & 70 & 70 & 65
      & \underline{65} & 60 & 45
      & 50 & 40 & 45
      & 20 & 20 & \underline{15}
      & \textbf{20}
      & 45 \\

    \rowcolor{light_mintgreen}Im2Flow2Act %
      & \checkmark
      & \checkmark
      & \underline{85} & \underline{80} & \textbf{85}
      & \textbf{75} & \underline{70} & \underline{60}
      & \underline{60} & \underline{55} & \textbf{55}
      & \underline{25} & \underline{25} & \textbf{20}
      & \underline{15}
      & \underline{55} \\

    \midrule

    \multicolumn{17}{l}{\textit{Goal-conditioned}} \\
    \midrule

    DP-Goal
      & 
      & 
      & 40 & 30 & 45
      & 50 & 25 & 30
      & 5 & 15 & 30
      & 5 & 5 & 0
      & 5
      & 22 \\

    Track2Act~\cite{track2act-bharadhwaj24eccv}
      & \checkmark
      & 
      & \underline{85} & \underline{80} & 75
      & 60 & 65 & 55
      & 55 & \underline{55} & 35
      & 15 & 15 & \textbf{20}
      & 10
      & 48 \\

    \rowcolor{mintgreen}\textbf{Ours}
      & \checkmark
      & \checkmark
      & \textbf{90} & \textbf{90} & \underline{80}
      & \textbf{75} & \textbf{75} & \textbf{70}
      & \textbf{65} & \textbf{65} & \underline{50}
      & \textbf{30} & \textbf{30} & \textbf{20}
      & \underline{15}
      & \textbf{58} \\
      
    \midrule

    Oracle
      & \checkmark
      & --
      & 95 & 90 & 90
      & 85 & 80 & 80
      & 75 & 70 & 55
      & 40 & 35 & 25
      & 20
      & 65 \\

    \bottomrule
  \end{tabular}
  }
  \vspace{-5mm}
  \label{tab:real-world-results}
\end{table}

%% file: tab/6-physical-qualitative.tex
\begin{figure*}[t]
    \centering
    \includegraphics[width=0.98\linewidth]{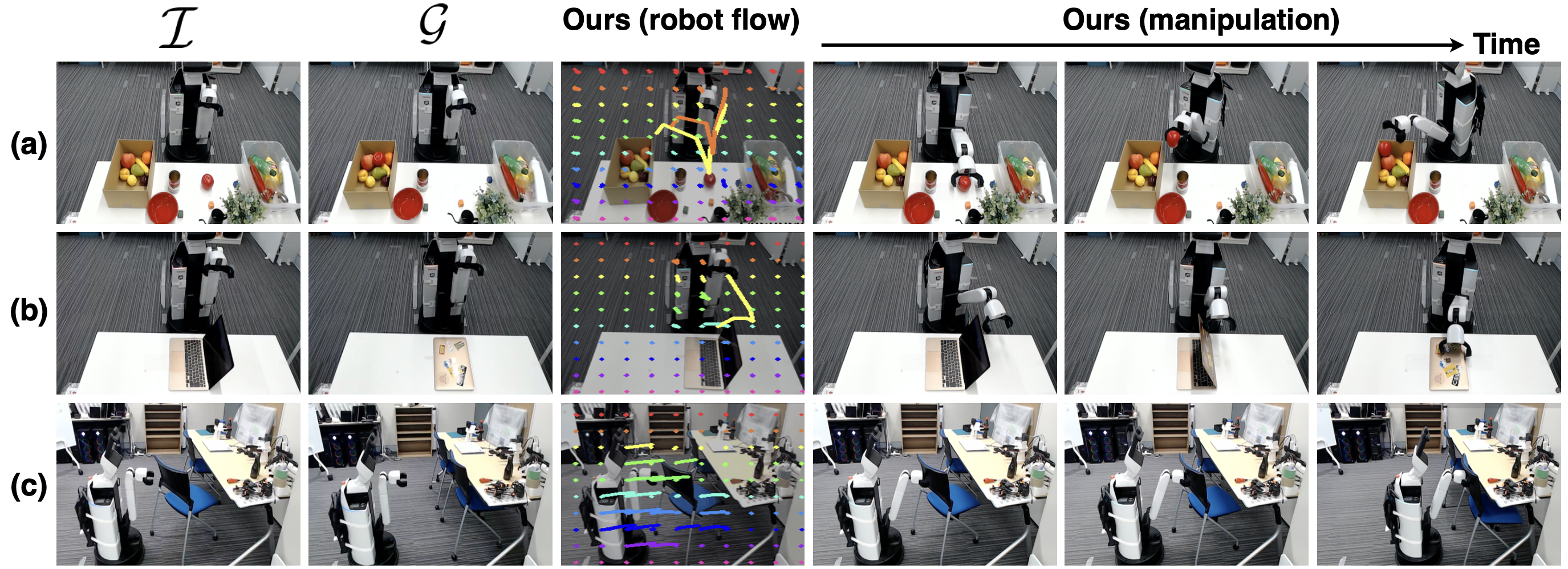}
    \vspace{-1mm}
    \caption{Qualitative results of our method in real-world experiments, showing three successful rollouts: (a) \textbf{bussing table}, (b) \textbf{close laptop}, and (c) \textbf{push chair}. The two left columns show $\mathcal{I}$ and $\mathcal{G}$, and the third column represents the robot flow generated by our method. The other columns show that the robot executed the manipulation conditioned on the flow.}
    \label{fig:real-world-qualitative}
    \vspace{-4mm}
\end{figure*}

%% file: section7.tex
\vspace{-2mm}
\section{Conclusion}
\vspace{-1mm}
In this study, we focused on flow-based object manipulation. 
The contributions of this study are as follows: We proposed \proposed, a framework that models robot velocity fields as probability velocity fields, which achieved the efficient and high-quality generation of \flows. In the flow generation task, our method outperformed the baseline methods on Fractal, Bridge V2, DROID-100, and Fanuc Manipulation datasets including zero-shot settings, while achieving approximately $24\times$ faster generation than standard flow matching. 
Moreover, through extensive real-world experiments with 260 trials per method across 13 mobile manipulation tasks, our method achieved a higher average success rate than the baseline methods.

\vspace{-2mm}
\section*{Limitations}
\vspace{-1mm}
Although our method achieved promising results, three primary limitations remain.
First, 2D \flows do not explicitly model scene depth or end-effector orientation. Thus, they can be insufficient for depth- and rotation-sensitive manipulation, which is typically required in realistic scenarios.
A possible solution to address this limitation is to extend the flow representation to incorporate geometry and rotation, including 3D or 6-DoF flows.
Second, our method assumes a static camera setup. When the camera moves, global motion can dominate the frame and obscure task-relevant flows. This global motion limits the use of egocentric human videos 
where camera shake is common, and complicates the implementation for mobile manipulation.
Potential directions include estimating camera parameters alongside flows and 
treating foreground and background motion separately.
Finally, while our formulation can represent multimodal per-timestep flow distributions, matching these marginals may lead to mode mixing across timesteps, as analyzed in~\cite{sfp-jiang25corl}.

%% file: supplymentaly.tex
\renewcommand{\thefigure}{\arabic{figure}}  
\setcounter{figure}{5}

\renewcommand{\thetable}{\arabic{table}} 
\setcounter{table}{3}

\renewcommand{\theequation}{\arabic{equation}}
\setcounter{equation}{7}
\section{Additional Related Work}
Robotic foundation models have become an active area of research across diverse tasks and environments~\cite{pi0-black25rss, pi05-black25corl, gr00t-bjorck25, giga-brain-ye25}.
Kawaharazuka et al.~\cite{survey-vla-kawaharazuka24ar} and Firoozi et al.~\cite{survey-robotic-foundation-firoozi25ijrr} provided complementary overviews of foundation models in robot learning, covering strategies for integrating multimodal inputs into robot control, architectural design choices, and benchmark datasets.
Moreover, generative modeling is another important factor for improving manipulation policies, and Urain et al.~\cite{survey-generative-for-robot-learning-urain26-tro} surveyed deep generative approaches for robot learning, including training objectives such as diffusion models~\cite{ddpm-ho20neurips}
and flow matching~\cite{flowmatching-lipman23iclr}.

\textbf{Human Datasets for Robot Learning.} 
Human videos provide scalable training data for robot learning, spanning internet-scale datasets obtained from the web and datasets intentionally recorded for embodied activity understanding~\cite{internvid-wang24iclr, webvid-10m-bain21iccv, 100-days-of-hands-shan20cvpr}. 
Internet-scale datasets such as WebVid-2M~\cite{webvid-10m-bain21iccv} and HD-VILA-100M~\cite{hd-vila-100m-xue22cvpr} are collected at scale and paired with automatically obtained text (\eg alt-text or transcripts), which provides broad coverage of human activities and object motions that can be useful for robot manipulation~\cite{hd-vila-100m-xue22cvpr,merlot-zellers21neurips, webvid-10m-bain21iccv, 100-days-of-hands-shan20cvpr}.
By contrast, Ego-4D~\cite{ego4d-grauman22cvpr},
Ego-Exo-4D~\cite{ego-exo-grauman24cvpr}, and EPIC-KITCHENS~\cite{epickitchens-damen18eccv}
are designed to capture egocentric human activities that involve object manipulation and provide temporally extended recordings of everyday tasks with free-form language annotations.
In practice, EPIC-KITCHENS and Something-Something-v2~\cite{sthsthv2-goyal17iccv}
serve as standard training datasets for robot learning approaches that leverage human videos~\cite{affordance-from-human-video-bahl23cvpr, adapt2reward-yang24eccv}.
EPIC-KITCHENS provides long-horizon egocentric videos of everyday tasks in real kitchens across many participants.
It has been used in several studies to learn visual features and temporally extended manipulation dynamics (\eg{}~\cite{affordance-from-human-video-bahl23cvpr}).
Something-Something-v2 consists of crowd-sourced short clips paired with language action templates.
The templates are determined primarily by how objects move and relate spatially.
Something-Something-v2 is used to learn motion-aware representations for downstream robot control and reward learning (\eg{}~\cite{adapt2reward-yang24eccv}).

\section{Dataset Details}\label{app:dataset}
We leveraged diverse videos available on the web, including both human and robot manipulation domains to train \firstmodule.
Specifically, we used human video clips from Something-Something-v2~\cite{sthsthv2-goyal17iccv}
and EPIC-KITCHENS~\cite{epickitchens-damen18eccv}.
Something-Something-v2 contains short human videos of everyday activities, and EPIC-KITCHENS comprises egocentric videos of people performing a variety of daily tasks in diverse kitchen environments.
We also used robot video clips from standard robot manipulation datasets Fractal~\cite{rt1-brohan23rss}
and Bridge V2~\cite{bridgev2-walke23corl}
for training.
For evaluation, we used Fractal and Bridge V2 as in-domain benchmarks, and further assessed zero-shot performance on DROID-100~\cite{droid-khazatsky24rss}
and Fanuc Manipulation~\cite{fanuc-zhu23},
which are also standard benchmarks in robot learning~\cite{oxe-neill24icra, openvla-kim24corl, univla-bu25rss}.

\tabref{tab:statistics} shows the statistics of the datasets used in our experiments.
The robot datasets were collected from multiple real-world robotic platforms, which resulted in variation in both the recorded motion data and the visual appearance of the scenes (\eg backgrounds and viewpoints). 
We used the training sets to train the model, the validation sets to tune the hyperparameters, and the test sets to evaluate the model.

\subsection{Preprocessing Details
\label{sec:preprocess}}

We curated the datasets as follows.
The human videos were automatically segmented into short clips of approximately 1--2 seconds that captured atomic object manipulations, following VITRA~\cite{vitra-li26icra}.
For each dataset, we defined $\mathcal{I}$ and $\mathcal{G}$ as the first and last frames of each video clip, respectively.
Each frame of the robot episodes was provided with a binary success label; for successful episodes, we defined $\mathcal{G}$ as the frame at which success was determined.
To obtain $\bm{\Xi}_{0:H-1}$, we used CoTracker3~\cite{cotracker3-karaev25iccv},
an off-the-shelf point-tracking method.
Specifically, we initialized a uniform grid of points (\eg 10 $\times$ 10) on $\mathcal{I}$ and tracked their trajectories throughout the clip.

Subsequently, we preprocessed the datasets as follows.
First, each image was resized to 256 $\times$ 256.
Next, we obtained a segmentation mask of either a human hand or a robot end-effector from $\mathcal{I}$ and incorporated it as an alpha channel of $\mathcal{I}$ in the model input.
In the case of human videos, hand masks were generated using SAM3~\cite{sam3-carion25} with the prompt ``human hand.''
When SAM3 failed to generate a hand mask, the corresponding sample was excluded from the dataset.
For robot videos, end-effector masks were generated using Robot-SAM~\cite{roboengine-yuan25iros}.
The pretrained model exhibited an over-segmentation issue: the generated masks covered entire robot arms rather than only the end-effectors. 
Therefore, we fine-tuned Robot-SAM to localize only end-effectors (see Sec.~\ref{sec:robot-sam}).

We also considered a dataset-specific issue in egocentric videos caused by camera motion.
In egocentric datasets such as EPIC-KITCHENS, videos were recorded using head-mounted cameras that move with head motion, and backgrounds typically dominate frames.
As a result, camera motion caused widespread global motion in the videos, which led to high flow magnitudes, where the flow magnitude is defined as the average motion vector magnitude over tracked points.
We considered clips with ground-truth flow magnitude exceeding 100 pixels as exhibiting substantial camera motion and excluded them from the datasets.

\input{tab/5-4-statistics}

\subsection{Details of Fine-Tuning Robot-SAM
\label{sec:robot-sam}}
We fine-tuned Robot-SAM~\cite{roboengine-yuan25iros} to obtain end-effector masks for robot videos, as described in Sec.~\ref{sec:preprocess}. 
Fig.~\ref{fig:sam} shows a qualitative comparison among SAM3~\cite{sam3-carion25}, Robot-SAM without fine-tuning, and Robot-SAM with fine-tuning. 
When we provided a prompt ``end-effector,'' SAM3 segmented the visible region of the robot arm rather than the end-effector itself. 
This result motivated the use of a robot-specific segmentation model. 
However, the pretrained Robot-SAM, a model specialized for robot-part segmentation, generated a mask covering the entire visible robot body, including both the end-effector and the arm. 
Therefore, to better localize end-effector regions, we fine-tuned Robot-SAM using 300 samples from the RoboSeg dataset~\cite{roboengine-yuan25iros} that we additionally annotated with end-effector masks. 
After fine-tuning, the model successfully localized the end-effector region while suppressing irrelevant arm regions.
\input{tab/supp-sam}

\section{Implementation Details}\label{app:implementation}
We used FLIP~\cite{flip-gao25iclr},
Im2Flow2Act~\cite{im2flow2act-xu24corl},
and GigaWorld-0-Video~\cite{giga-world-0-ye25}
as language-conditioned baseline methods, and Track2Act~\cite{track2act-bharadhwaj24eccv}
as a goal-conditioned baseline method.
We selected FLIP and Im2Flow2Act because of their strong capabilities in language-conditioned \flow generation and trained them on the same datasets used for training our method.
Additionally, we included GigaWorld-0-Video, a language-conditioned world model specialized for robotic environments, to assess the applicability of a world model to the flow generation task.
For GigaWorld-0-Video, a video was generated from $\mathcal{I}$ and a language input, and the \flow was subsequently obtained using CoTracker3~\cite{cotracker3-karaev25iccv}.
Language instructions were used as inputs for each language-conditioned method; instructions included in the original datasets were used directly, and otherwise they were generated from $\mathcal{I}$ and $\mathcal{G}$ using Qwen2.5-VL~\cite{qwen2.5-vl-bai25}.
Track2Act was selected as the primary baseline method because it has demonstrated competitive performance under the same goal-conditioned setting as our method.

\tabref{tab:params} shows the experimental settings of our method.
Following previous work~\cite{track2act-bharadhwaj24eccv},
we set $H = 8$.
We trained our method using two NVIDIA H200 SXM GPUs (VRAM 141 GB), and used a single H200 for inference.
The total training time was approximately 19 hours.
Our method had approximately 480M trainable parameters and 2.2G multiply-add operations.
\input{tab/5-8-params}
\section{Evaluation Metrics Details}\label{app:evaluation}

In Sec.~\ref{sec:experiments}, we used ADE, FDE, and LTDR as evaluation metrics.
ADE was computed by averaging the Euclidean distance between $\mathcal{X}_{0:H-1}$ and $\bm{\Xi}_{0:H-1}$ over every point and timestep.
FDE was calculated by averaging the Euclidean distance between $\bm{X}_{H-1}$ and $\bm{\Xi}_{H-1}$ over every point at the final step.
LTDR followed a delta-based evaluation protocol~\cite{tapvid-doersch22neurips}.
For each step $h$ and threshold $m$, $\delta_m^h$ denotes the fraction of points for which the Euclidean distance between $\mathcal{X}_h$ and $\bm{\Xi}_h$ was within $m$ pixels.
The final score was computed by averaging $\delta_m^h$ over thresholds and the flow horizon as:
\begin{align}
\text{LTDR} = \frac{1}{MH} \sum_{m=1}^{M} \sum_{h=0}^{H-1} \delta_m^h .
\end{align}
The LTDR score is in $[0,1]$, where higher values indicate better performance, and we set $M=100$ in our experiments.

These metrics are typically computed over the full keypoint set; however, we restricted the evaluation to end-effector regions. This prevents background-dominated evaluation and better reflects end-effector robot-flow quality.
In practice, under a standard fixed-camera setup, a large fraction of keypoints tends to lie on stationary background regions. Therefore, when evaluating the full keypoint set, the resulting scores are often dominated by such task-irrelevant stationary regions.
As a result, even degenerate outputs such as keeping every keypoint fixed (a static prediction) can result in misleadingly favorable scores despite containing no task-relevant motion.

To mitigate this issue, we extracted the end-effector regions under the assumption that points corresponding to the end-effector exhibit larger displacement than those in other regions.
We obtained these regions by identifying a cluster of points in the vicinity of the point with the maximum displacement, followed by displacement-based thresholding and morphological operations using dilation and erosion.

\figref{fig:static_qualitative} shows representative cases from Fractal~\cite{rt1-brohan23rss}
and Bridge V2~\cite{bridgev2-walke23corl}
in which evaluating the full keypoint set was problematic. In both cases, the predicted robot flows qualitatively represented appropriate end-effector motions. Specifically, the generated flow in \figref{fig:static_qualitative}(i) captured the end-effector motion that moved the can toward the back of the table, and in \figref{fig:static_qualitative}(ii), the flow corresponded to placing the blue object onto the towel. However, when the full keypoint set was evaluated, the predictions of our method yielded higher ADE scores than the static predictions.

\input{tab/supp-static-quantitative}
Moreover, \tabref{tab:static_quantitative} shows the quantitative comparison of ADE scores between the static predictions and our method on the Bridge V2 test set. When the full keypoint set was evaluated, the static prediction resulted in an ADE of 4.59, which was misleadingly lower than our method's ADE of 4.84. Therefore, we restricted the evaluated keypoints to end-effector regions.
Under this setting, the metric better reflected manipulation-relevant motion quality: our method achieved an ADE of 27.11, which is substantially lower than the ADE of 49.32 obtained by the static prediction.

\input{tab/supp-static-qualitative}

\section{Additional Results of Flow Generation}

\input{tab/supp-ablation}
\subsection{Additional Ablation Studies}
\tabref{tab:supp-ablation} shows additional ablation studies on the key components of our method. We evaluated the effects of the target velocity field design and the use of human videos.

\noindent\textbf{Target velocity field ablation}. To assess the contributions of our formulation (Eq.~(4)), we compared the full formulation (Setting (iv)) with two ablated settings: \textit{No feedback} (Setting (i)), where the stabilizing feedback term was removed ($k=0$), and \textit{Sparse} (Setting (ii)), where the sampled coordinates were not perturbed around the target points ($\sigma_0=0$).
In Setting (ii), the sampled coordinates $X$ were identical to $\Xi_h$.
Thus, the model learned velocities only at sparse keypoints, rather than modeling robot flows as dense velocity fields.
The results show that both settings degraded the ADE scores across the datasets compared with the full formulation.
Specifically, Setting (i) degraded the ADE scores by 13.95, 15.20, 3.35, and 7.08 points on Fractal, Bridge V2, DROID-100, and Fanuc Manipulation, respectively. Similarly, Setting (ii) increased the ADE scores by 14.34, 15.46, 3.45, and 8.67 points, respectively. These results indicate the effectiveness of both stabilizing feedback and dense velocity-field modeling for high-quality flow generation.

\noindent\textbf{Human video ablation.}
To evaluate the effectiveness of human-video training, we compared a model trained only on the robot datasets (Setting (iii)) with our full model (Setting (iv)). 
Setting (iii) yielded ADE scores of 28.58, 33.67, 36.42, and 31.22 on Fractal, Bridge V2, DROID-100, and Fanuc Manipulation, respectively. 
Compared with Setting (iv), Setting (iii) degraded the ADE scores by 7.35, 6.56, 0.53, and 8.76 points, respectively. 
These results indicate that including human videos in training improves robot-flow generation across both in-domain and zero-shot settings.

\subsection{Additional Qualitative Result}
Fig.~\ref{fig:flow-failure} shows a failure case on Fanuc Manipulation.
The four left columns show sampled frames from $\mathcal{I}$ to $\mathcal{G}$ in temporal order, and the remaining columns show the robot flows of the ground truth, Track2Act, and our method.
In this sample, the robot grasped the blue object located at the bottom of the image and placed it into the basket; both methods failed to generate appropriate \flows.
Track2Act failed to generate a flow corresponding to the end-effector motion.
Our method generated the flow toward the correct object, but it represented only the lifting motion and did not reach the basket.
We hypothesize that this failure was caused by the absence of the manipulated object in $\mathcal{G}$, which made the object's destination unclear.
This result indicates that, although our method captured object-level correspondence between $\mathcal{I}$ and $\mathcal{G}$, it struggled to capture the semantic properties of objects in the scene (e.g., a basket serves as a container).

\input{tab/6-9-qualitative}

\section{Error Analysis of Flow Generation}

To investigate the limitations of our method, we conducted an error analysis of $100$ failure cases from Bridge V2.
We define failure cases as samples for which our method exhibited higher ADE scores than a baseline method, Track2Act.  
There were $141$, $249$, $33$, and $114$ failure cases in Fractal, Bridge V2, DROID-100, and Fanuc Manipulation, respectively.
Table~\ref{tab:error-analysis} shows the results of the error analysis.
We identified five major failure modes:

\noindent\textbf{Intermediate trajectory deviation:}
This category refers to cases in which the generated \flows represented the manipulation of the correct object and reached the appropriate target location, but the intermediate trajectories exhibited significant discrepancies from the ground-truth flows{}. A representative example is when a source demonstration depicted the end-effector picking an apple from the left and moving it to the right. Although the generated flow correctly targeted the apple and the goal location, the intermediate flow deviated substantially from the ground-truth trajectory.

\noindent\textbf{Inappropriate behavior in source demonstrations:} 
This category includes cases in which robot manipulation in the corresponding source demonstration was unsuitable for the task. In Bridge V2, some trajectories with little or no interaction with objects were intentionally included~\cite{bridgev2-walke23corl}.
In such cases, unchanged object configurations between $\mathcal{I}$ and $\mathcal{G}$ did not provide substantial task-relevant constraints on the end-effector trajectories. 

\noindent\textbf{Target object comprehension error:} This category refers to cases in which the generated \flows represented the manipulation of incorrect objects. These failures often stemmed from challenging visual conditions, such as target occlusion in $\mathcal{I}$ or $\mathcal{G}$, or small displacements.

\noindent\textbf{Target location mismatch:}
This category includes cases in which the errors between the generated flows and the ground-truth flows were small at intermediate timesteps but became significant at the terminal positions. A representative example is when the generated flow represented the manipulation of the correct target object and reflected the appropriate direction, but failed to reach the precise final position.

\noindent\textbf{Tracking error in ground-truth flows:}
This category refers to cases in which the ground-truth \flows were unreliable because of the tracking errors of CoTracker3~\cite{cotracker3-karaev25iccv}.
A typical case was when the end-effector moved out of the frame, which led to the failure to track the end-effector trajectory properly.

Table \ref{tab:error-analysis} shows intermediate trajectory deviation as the primary bottleneck.
A possible solution for this is to incorporate action chunking~\cite{act-zhao23rss} to simultaneously output \flows over multiple consecutive timesteps.
This approach enables the model to explicitly learn temporal consistency, which is expected to facilitate the generation of \flows that are better aligned with robot motions.

\input{tab/6-13-error-analysis}

\section{Discussion on Efficient Flow Generation in Flow as Flow}

As shown in Table 1, Flow as Flow substantially accelerated robot-flow generation, particularly compared with diffusion-based methods (e.g., 33$\times$ faster than Track2Act~\cite{track2act-bharadhwaj24eccv}
and 24$\times$ faster than Im2Flow2Act~\cite{im2flow2act-xu24corl}).
Fig.~\ref{fig:supp-efficiency} illustrates the conceptual difference between conventional diffusion-based generation and Flow as Flow.
Diffusion models demonstrate strong generative capacity; however, they typically require many sampling steps at inference time.
This bottleneck also appears in representative diffusion-based flow generation methods (e.g., Track2Act uses 250 diffusion steps).

In contrast, as formulated in Eq.~(6), Flow as Flow generates robot flows in only $H$ steps (e.g., 8 steps).
In typical flow-based manipulation settings, the flow horizon $H$ is substantially smaller than the number of sampling steps used in diffusion-based generation~\cite{track2act-bharadhwaj24eccv, im2flow2act-xu24corl}.
As a result, Flow as Flow reduces the number of model evaluations at inference time.

\input{tab/supp-efficiency}

\section{Details of Real-World Experiments}

\subsection{Task Details}\label{app:task}
In the real-world experiments, we evaluated our method on a broad set of 13 mobile manipulation tasks (Fig.~4).
The detailed task descriptions are as follows:
 
\textbf{Bin picking}: Clear the specified object from a table by grasping it, rotating the mobile base, and placing it into a bin.

\textbf{Bussing table}: Grasp the specified object on a table and place it into the designated receptacle, either a bin or a cardboard box, after rotating the mobile base. Compared with ``bin picking,'' this task is more complex because it requires selecting the appropriate destination among multiple possible receptacles.

\textbf{Push bin into shelf}: Push a drawer-like storage bin back into a shelf by moving the mobile base forward.

\textbf{Push chair}: Push a chair under a table. This task involves a motion pattern similar to ``push bin into shelf,'' but is more demanding because the slightly curved backrest makes it difficult to maintain a straight pushing direction.

\textbf{Open drawer}: Grasp the small handle of a tabletop drawer and pull the drawer open by moving the mobile base backward. This task requires the precise grasping of a small handle and controlled pulling in an appropriate direction.

\textbf{Close drawer}: Close an open tabletop drawer by pushing it forward.

\textbf{Put fruit on plate}: Grasp the specified fruit, rotate the mobile base, and place the fruit on a plate without colliding with other objects already on the plate.

\textbf{Close box}: Close the lid of a cardboard box. This task requires inserting the end-effector under the lid according to the initial opening angle of the box.

\textbf{Water plant}: Grasp a watering can, rotate the mobile base toward the plant, and water the plant.

\textbf{Take towel}: Remove a towel hanging on a mug stand while avoiding snagging the towel on the stand.

\textbf{Close laptop}: Close an open laptop by manipulating its articulated screen. Successful execution requires adapting to different initial opening angles of the laptop.

\textbf{Stack block}: Rotate the mobile base and stack a block on top of stacked blocks. This task is particularly challenging because it requires grasping a small block and precisely aligning it with the stacked target blocks; even a few degrees of base-rotation error can cause misalignment and lead to failure.

\textbf{Stack cup}: Rotate the mobile base and stack a cup onto another cup. Similar to ``stack block,'' this task is sensitive to base motion because a small base-rotation error of only a few degrees can result in failure.

For language-conditioned methods, we used the instructions ``Put \(\langle object \rangle\) in the bin,'' ``Put \(\langle object \rangle\) in \(\langle receptacle \rangle\),'' ``Close drawer,'' ``Put the chair away,'' ``Open top drawer,'' ``Close top drawer,'' ``Put \(\langle fruit \rangle\) on the plate,'' ``Close the cardboard box,'' ``Pour water onto the plant,'' ``Take the towel from the rack,'' ``Close the laptop,'' ``Stack block,'' and ``Stack cup,'' respectively.
\(\langle object \rangle\), \(\langle fruit \rangle\), and \(\langle receptacle \rangle\) denote the names of the target object, fruit, and receptacle, respectively, where \(\langle receptacle \rangle\)  is either ``bin'' or ``cardboard box.''

\subsection{Additional Qualitative Results}
\figref{fig:supp-real-world-qual} shows additional qualitative results from the real-world experiments, including the cases of \textbf{bin picking}, \textbf{close box}, \textbf{close drawer}, \textbf{push bin into shelf}, \textbf{stack cup}, \textbf{take towel}, and \textbf{water plant}.
In each case, our method generated an appropriate robot flow, and the robot successfully executed object manipulation conditioned on the generated flow. Specifically,~\figref{fig:supp-real-world-qual}(e) shows a \textbf{stack cup} case. This task is sensitive to the robot's base pose: even a small positional offset can prevent the cup from fitting onto the other cup.
Despite this difficulty, the robot properly rotated its base and successfully stacked the cup. 
Moreover,~\figref{fig:supp-real-world-qual}(f) illustrates a \textbf{take towel} case.
In this case, the robot appropriately grasped a deformable object.

\section{Discussion on Effectiveness of Flow-Based Approaches}\label{app:flow-effectiveness}

Table~3 shows that the flow-based methods generally achieved higher average success rates than both non-flow-based methods (DP-Lang and DP-Goal).
We hypothesize that this difference arises from the distinct characteristics of each conditioning modality, as detailed below.

Although goal images provide rich information about desired object states and spatial arrangements, they also contain substantial task-irrelevant visual information (e.g., background appearance), which can vary across environments.
Therefore, they may lead to policies that are sensitive to appearance changes.
By contrast, language instructions are largely invariant to environmental conditions, and thus provide more stable conditioning signals than goal images.
However, they are typically limited to short and coarse task descriptions (e.g., ``Place the dishes in the sink.'')~\cite{pi05-black25corl},
and often lack fine-grained motion details.

By contrast, robot flows can address these limitations by providing task-relevant motion representations as effective conditioning signals.
Although FLIP is also a flow-based method, its lower performance may be caused by its relatively small model size (55M, compared with 120M for DP-Lang) and lower-capacity autoregressive point-coordinate prediction, which may limit its ability to leverage pretraining in our real-world domain.

\input{tab/supp-real-world-qual}

%% file: tab/5-4-statistics.tex
\begin{table*}[t]
\centering
\caption{Statistics of the datasets used in our experiments. 
The original DROID-100 videos were recorded at 15 fps and downsampled to 5 fps for consistency with the frame rates of the other robot datasets.}
\resizebox{\linewidth}{!}{
\begin{tabular}{lclcc}
\hline
Dataset
& Embodiment
& \# Samples (train / val. / test)
& Avg. frame rate [fps]
& Avg. duration [s] \\
\hline
Something-Something-v2~\cite{sthsthv2-goyal17iccv}
& Human
& \makebox[\widthof{87,212}][r]{49,257} (46,257 / 3,000 / -)
& 12.00 & 1.78  \\
EPIC-KITCHENS~\cite{epickitchens-damen18eccv}
& Human
& \makebox[\widthof{87,212}][r]{54,938} (51,938 / 3,000 / -)
& 56.09 & 0.83  \\
Fractal~\cite{rt1-brohan23rss}
& Google Robot
& \makebox[\widthof{87,212}][r]{87,212} (84,212 / 1,000 / 2,000)
& 3.00  & 14.15 \\
Bridge V2~\cite{bridgev2-walke23corl}
& WidowX
& \makebox[\widthof{87,212}][r]{60,064} (57,064 / 1,000 / 2,000)
& 5.00  & 7.51  \\
DROID-100~\cite{droid-khazatsky24rss}
& Franka
& \makebox[\widthof{87,212}][r]{100} (- / - / 100)
& 5.00  & 21.47 \\
Fanuc Manipulation~\cite{fanuc-zhu23}
& Fanuc Mate
& \makebox[\widthof{87,212}][r]{415} (- / - / 415)
& 5.00  & 30.17 \\
\hline
\end{tabular}
}
\vspace{-4mm}
\label{tab:statistics}
\end{table*}

%% file: tab/supp-sam.tex
\begin{figure*}[h]
    \centering
  \centering
  \includegraphics[width=0.90\linewidth]{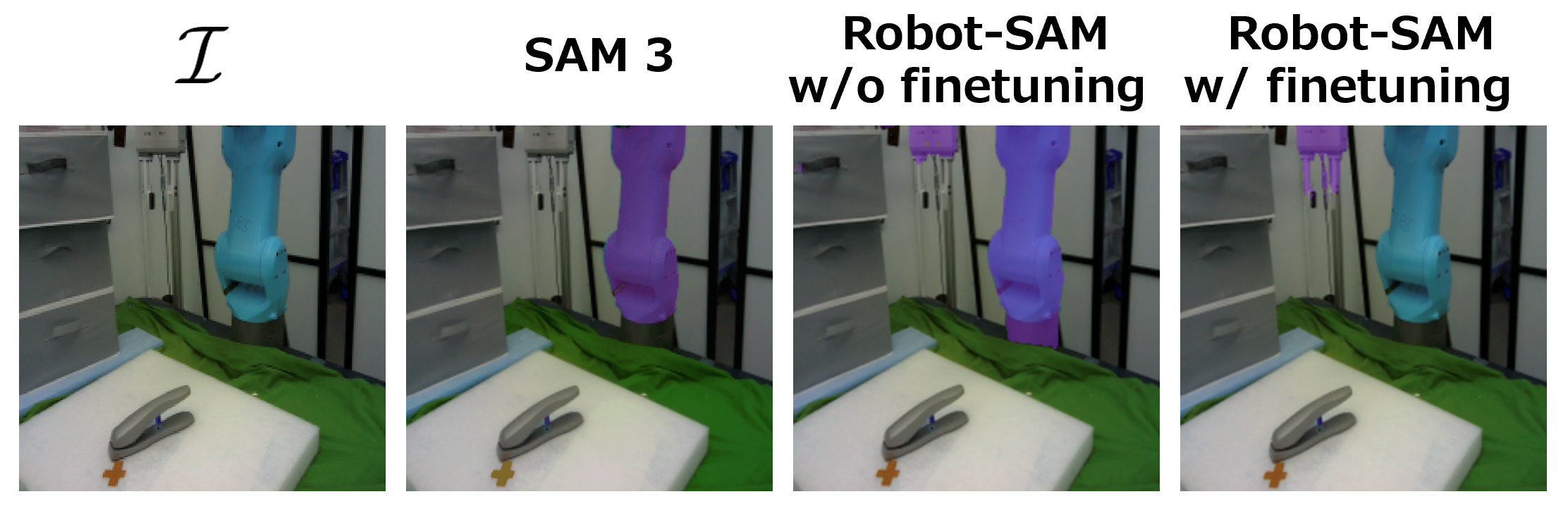}
  \caption{\textbf{Qualitative comparison of end-effector masks.}
The left image shows $\mathcal{I}$, and the other images present generated masks of SAM3~\cite{sam3-carion25}, Robot-SAM~\cite{roboengine-yuan25iros} without fine-tuning, and Robot-SAM with fine-tuning.
The generated segmentation masks are overlaid in purple.}
  \label{fig:sam}
\end{figure*}

%% file: tab/5-8-params.tex
\begin{table}[h]
\centering
\caption{Experimental setup.}
\begin{tabular}{lc}
\hline
Optimizer        & AdamW \\
Batch size       & 512 \\
Learning rate    & $1.0\times10^{-4}$ \\
LR schedule      & Cosine decay \\
Warmup steps     & 9,000 \\
Training steps   & 100,000 \\
Weight decay     & 0.01 \\
EMA decay        & 0.9999 \\
$H$              & 8 \\
$\sigma_0$       & 0.05 \\
$k$              & 1 \\
\hline
\end{tabular}
\label{tab:params}
\end{table}

%% file: tab/supp-static-quantitative.tex
\begin{wraptable}{r}{0.50\textwidth}
  \centering
  \small
  \setlength{\tabcolsep}{6pt}
  \renewcommand{\arraystretch}{1.15}
\caption{Quantitative comparison of ADE on the Bridge V2 test set for two evaluation regions: the full keypoint region and end-effector region. 
\textbf{Bold} denotes the best score.}
  \vspace{3pt} 
  \label{tab:static_quantitative}
  \begin{tabular}{ l c c }
    \toprule
    Method & Full keypoint set & End-effector only\\ 
    \midrule
    Static & \textbf{4.59} & 49.32 \\
    \textbf{Ours} & 4.84 & \textbf{27.11} \\
    \bottomrule
  \end{tabular}
  \vspace{15pt}
\end{wraptable}

%% file: tab/supp-static-qualitative.tex
\begin{figure*}[h]
    \centering
  \centering
  \includegraphics[width=\linewidth]{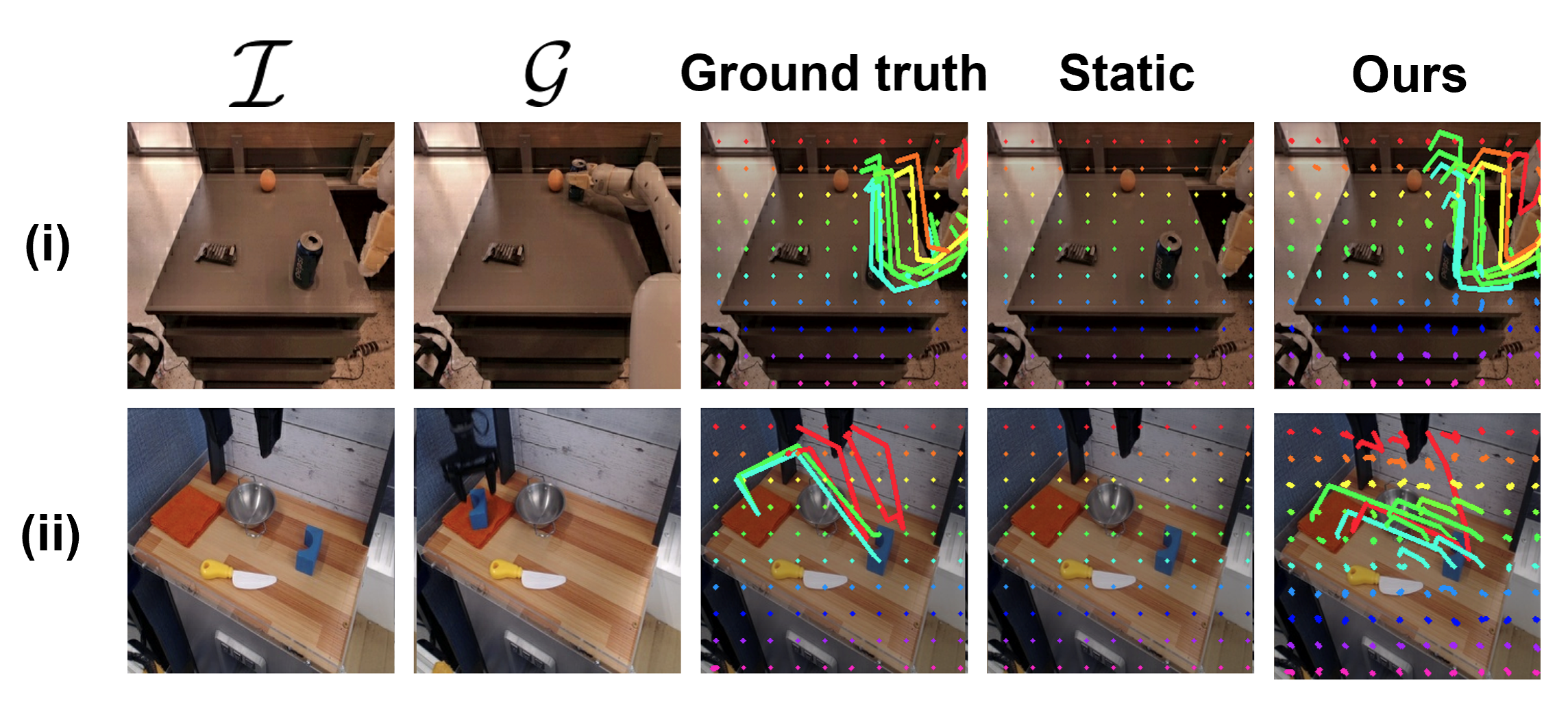}
  \caption{\textbf{Representative cases where evaluating the full keypoint set could be misleading.} The two left columns show $\mathcal{I}$ and $\mathcal{G}$. The other columns represent the ground truth flows, the static predictions, and the predictions of our method. The static predictions were obtained by keeping every keypoint fixed. Rows (i) and (ii) show samples from Fractal~\cite{rt1-brohan23rss}
  and Bridge V2~\cite{bridgev2-walke23corl},
  respectively. In both cases, the static predictions resulted in lower ADEs than the predictions of our method under full-keypoint evaluation: 3.50 vs. 4.27 for (i) and 2.26 vs. 4.45 for (ii).}
  \label{fig:static_qualitative}
\end{figure*}

%% file: tab/supp-ablation.tex
\begin{table*}[t]
  \centering
  \small
  \setlength{\tabcolsep}{4pt}
  \renewcommand{\arraystretch}{1.15}
  \caption{The ADE results of  additional ablation studies (the lower the better).}
 \resizebox{\linewidth}{!}{
  \setlength{\aboverulesep}{0pt}
  \setlength{\belowrulesep}{0pt}
  \begin{tabular}{
    l |
    c c c
    c c c
    c c c
    c c c
  }
    \toprule
    Setting & 
    Fractal~\cite{rt1-brohan23rss}
    &
    Bridge V2~\cite{bridgev2-walke23corl} 
    &
    DROID-100~\cite{droid-khazatsky24rss}
    &
    Fanuc Manipulation~\cite{fanuc-zhu23}
    \\
    \midrule

    (i)
    No feedback ($k=0$)
    & 35.18 
    & 42.31 
    & 39.24 
    & 29.54 \\
      
    (ii)
    Sparse ($\sigma_0=0$)
    & 35.57 
    & 42.57 
    & 39.34 
    & 31.13 \\

    (iii)
    w/o human videos
    & 28.58
    & 33.67
    & 36.42
    & 31.22 \\

    (iv) 
    \textbf{Ours}
      & \textbf{21.23}  
      & \textbf{27.11}  
      & \textbf{35.89}  
      & \textbf{22.46} \\

    \bottomrule
  \end{tabular}
  }
  \label{tab:supp-ablation}
\end{table*}

%% file: tab/6-9-qualitative.tex
\begin{figure*}[h]
    \centering
    \includegraphics[width=1.0\linewidth]{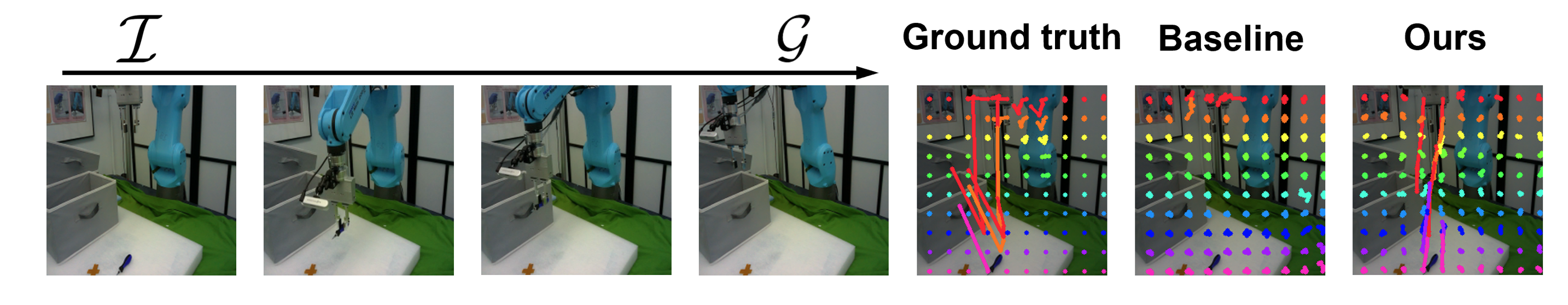}
    \caption{A failure case for Fanuc Manipulation. The four left columns show frames from $\mathcal{I}$ to $\mathcal{G}$. The remaining columns show the robot flows of the ground truth, a baseline method (Track2Act), and our method.}
    \label{fig:flow-failure}
\end{figure*}

%% file: tab/6-13-error-analysis.tex
\begin{table}[h]
  \centering
  \caption{Categorization of failure cases.}
  \renewcommand{\arraystretch}{1.3}
  \begin{tabular}{lc}
    \toprule
    Error category & \# Errors \\
    \midrule
    Intermediate trajectory deviation & 39 \\
    Inappropriate behavior in source demonstrations & 21 \\
    Target object comprehension error & 20 \\
    Target location mismatch & 16 \\
    Tracking error in ground-truth flows & 4 \\
    \midrule
    Total & 100 \\
    \bottomrule
  \end{tabular}
  \label{tab:error-analysis}
\end{table}

%% file: tab/supp-efficiency.tex
\begin{figure*}[h]
    \centering
  \centering
  \includegraphics[width=0.8\linewidth]{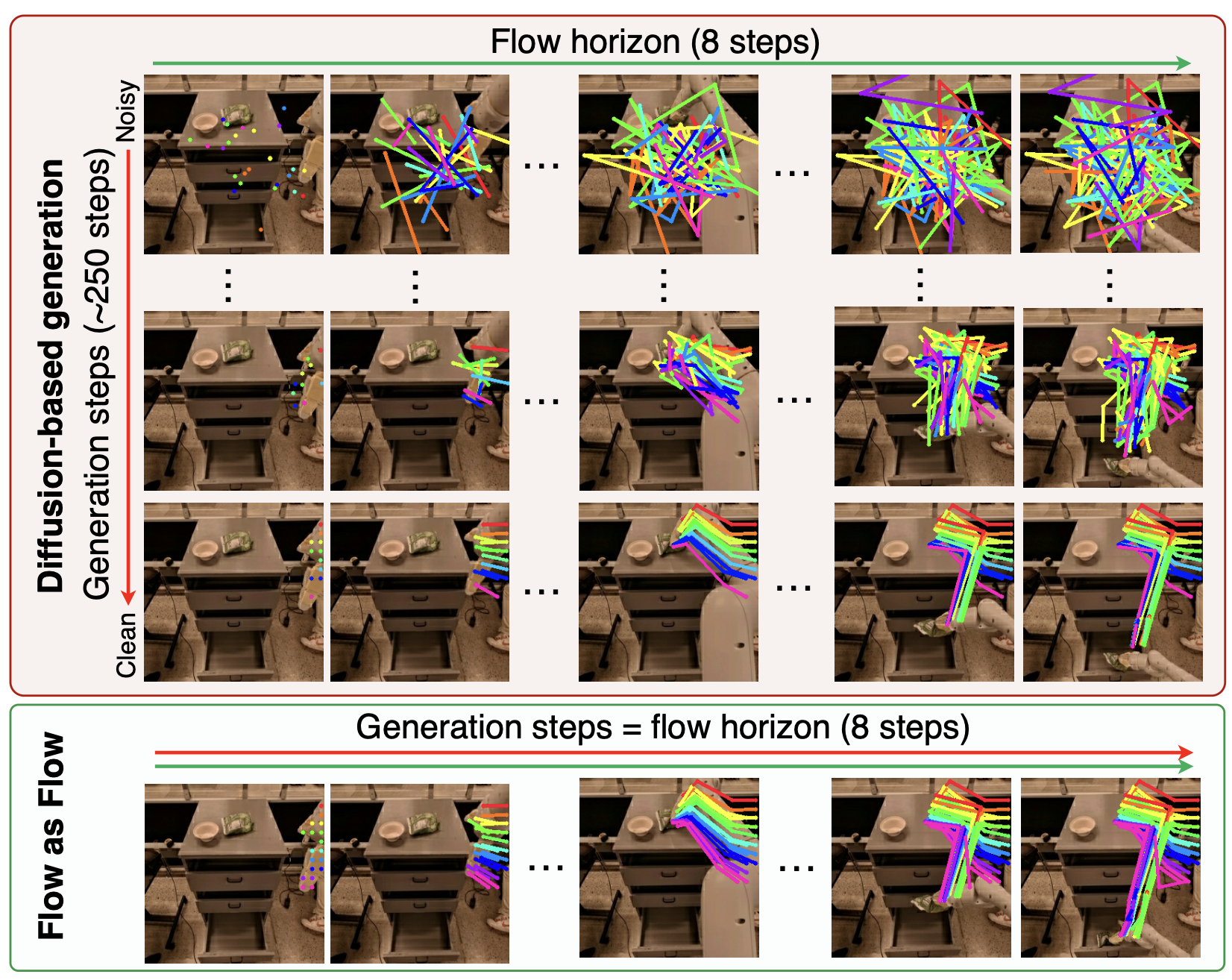}
  \caption{\textbf{Conceptual difference between conventional diffusion-based generation and Flow as Flow.}
Flow as Flow substantially accelerates diffusion-based flow generation methods by unifying generation steps with the flow horizon.}
  \label{fig:supp-efficiency}
\end{figure*}

%% file: tab/supp-real-world-qual.tex
\begin{figure*}[h]
    \centering
  \centering
  \includegraphics[width=0.95\linewidth]{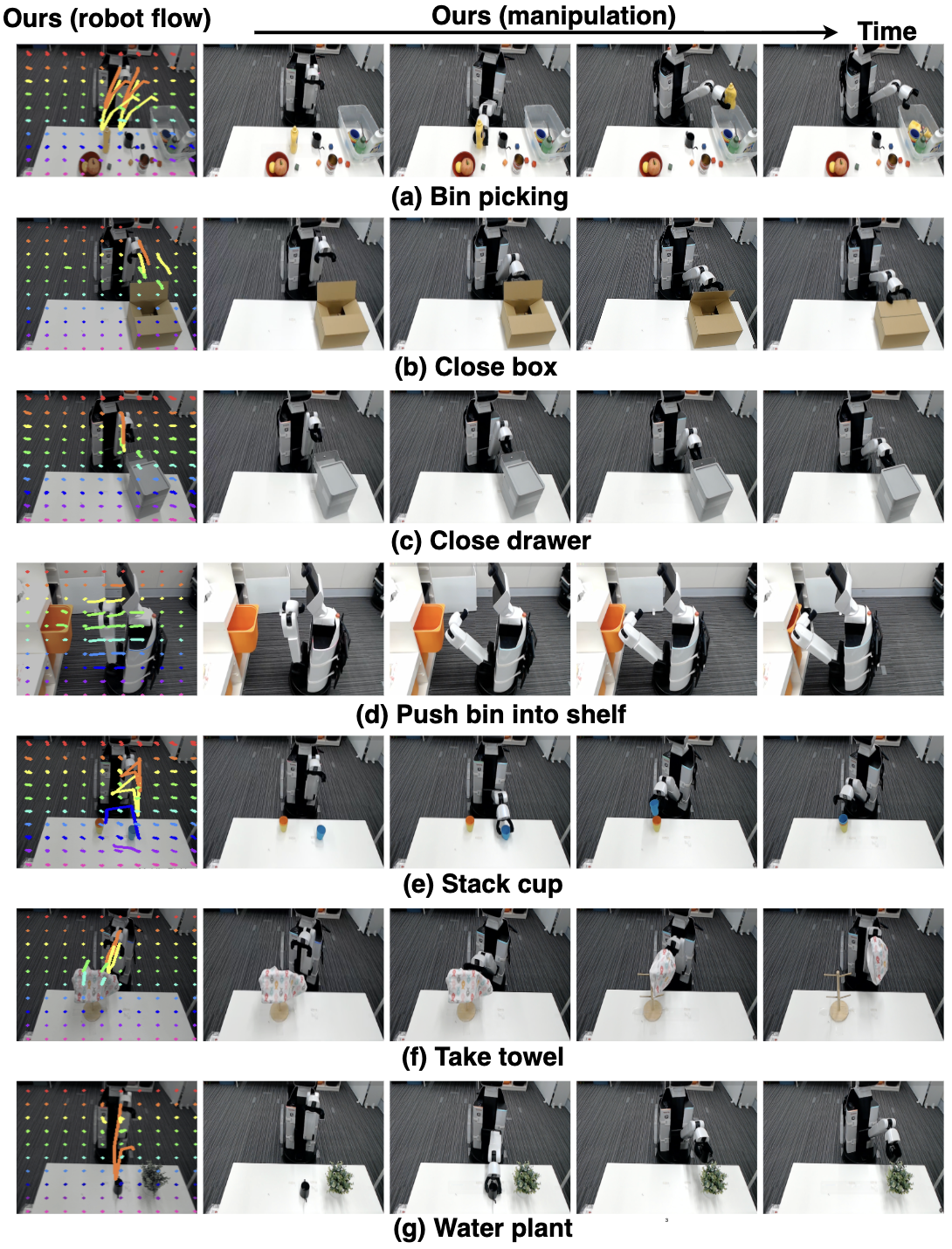}
  \caption{Additional qualitative results in the real-world experiments. The left column represents the robot flow generated by our method. The other columns show
that the robot executed the manipulation conditioned on the flow.}
  \label{fig:supp-real-world-qual}
\end{figure*}